%% file: article-arxiv.tex
\documentclass{article}

\usepackage{fontenc, inputenc, calc, indentfirst, fancyhdr, graphicx, epstopdf, lastpage, ifthen, lineno, float, amsmath, setspace, enumitem, mathpazo, booktabs, titlesec, etoolbox, tabto, xcolor, soul, multirow, microtype, tikz, totcount, changepage, attrib, upgreek, amsthm, hyphenat, hyperref, cleveref, footmisc, url, geometry, newfloat, caption}

\usetikzlibrary{fit,automata,arrows,positioning,calc,petri,topaths,arrows.meta}
\tikzstyle{nrect} = [rectangle, minimum size=7mm, thick, draw=black, node distance=5mm, rounded corners=0.1ex]
\definecolor{themeblue}{RGB}{25,71,163}
\tikzset{
>={Triangle[length=2.4mm,width=2.0mm]},
dedge/.style={arrows=->, black, thick},
}

\usepackage{comment}
\usepackage{multirow}
\usepackage{dblfloatfix}
\usepackage{xcolor}
\usepackage{caption}
\usepackage{subcaption}
\usepackage{pdflscape}
\def\startlandscape{\begin{landscape}}
\def\finishlandscape{\end{landscape}}

\input{math}
\renewcommand{\norm}[1]{\enVert{#1}}

\usepackage{listings}
\newcounter{nalg} 
\renewcommand{\thenalg}{\arabic{nalg}} 
\DeclareCaptionLabelFormat{algocaption}{Algorithm \thenalg} 

\lstnewenvironment{algorithm}[1][] 
{
	\refstepcounter{nalg} 
	\captionsetup{labelformat=algocaption, labelsep=period, textfont={small, stretch=1.17}, aboveskip=6pt, singlelinecheck=off, justification=justified}
	\definecolor{codecomment}{rgb}{0.5,0.5,0.5}
	\definecolor{codekw}{rgb}{0.1, 0.3, 0.6}
	\lstset{ 
		language=sh,
		columns=fullflexible,  
		mathescape=true,
		breaklines=true,
		frame=leftline,
		commentstyle=\color{codecomment},
		keywordstyle=\color{codekw}\bfseries,
		keywords={,procedure, input, output, return, datatype, function, in, if, else, for, foreach, while, with, begin, end, }
		xleftmargin=1em,  
		#1
	}
}
{}

\newcommand{\delete}[1]{}
\newcommand{\noarticle}[1]{}

\newcommand{\mytitle}{Revisiting Consistency for Semi-Supervised\\Semantic Segmentation}
\title{\mytitle\thanks{The journal version of this article is available at \url{https://www.mdpi.com/1424-8220/23/2/940}. This work has been supported by Croatian Science Foundation, grant IP-2020-02-5851 ADEPT. This work has also been supported by European Regional Development Fund, grant KK.01.1.1.01.0009 DATACROSS, and by VSITE College for Information Technologies.}}
\date{}

\author{
	Ivan Grubišić,
	Marin Oršić and
	Siniša Šegvić
}

\newcommand{\address}{{
  \footnotesize
  \begin{center}
  	\textit{University of Zagreb Faculty of Electrical Engineering and Computing, Unska 3, 10000 Zagreb\\
  	\{ivan.grubisic, sinisa.segvic\}@fer.hr\\
  	Microblink Ltd., Strojarska cesta 20, 10000 Zagreb\\
  	marin.orsic@microblink.com}
  \end{center}
	\bigskip
}}

\begin{document}
\maketitle
\address

\begin{abstract}
	Semi-supervised learning an attractive technique in  practical deployments of deep models since it relaxes the dependence on labeled data. It is especially important in the scope of dense prediction because pixel-level annotation requires significant effort. This paper considers semi-supervised algorithms that enforce consistent predictions over perturbed unlabeled inputs. We study the advantages of perturbing only one of the two model instances and preventing the backward pass through the unperturbed instance. We also propose a competitive perturbation model as a composition of geometric warp and photometric jittering. We experiment with efficient models due to their importance for real-time and low-power applications. Our experiments show clear advantages of (1) one-way consistency, (2) perturbing only the student branch, and (3) strong photometric and geometric perturbations. Our perturbation model outperforms recent work and most of the contribution comes from photometric component. Experiments with additional data from the large coarsely annotated subset of Cityscapes suggest that semi-supervised training can outperform supervised training with the coarse labels. Our source code is available at \url{https://github.com/Ivan1248/semisup-seg-efficient}.  
\end{abstract}




\section{Introduction}

Most machine learning applications
are hampered by the need
to collect large annotated datasets.
Learning with incomplete supervision
\cite{kolesnikov19cvpr,lee19cvpr}
presents a great opportunity
to speed up the development cycle
and enable rapid adaptation
to new environments. 
Semi-supervised learning
\cite{tarvainen17nips,miyato19pami,xie20nips}
is especially relevant
in the dense prediction context
\cite{souly17iccv,hung18bmvc,mittal19pami}
since pixel-level labels are very expensive,
whereas unlabeled images are easily obtained.

Dense prediction typically operates on high resolutions
in order to be able to recognize small objects.
Furthermore, competitive performance
requires learning on large batches and large crops
\cite{cordts15cvprw,neuhold17iccv,maggiori17igarss}.
This typically entails
a large memory footprint during training,
which constrains model capacity \cite{bulo18cvpr}.
Many semi-supervised algorithms introduce additional components
to the training setup.
For instance, training
with surrogate classes \cite{dosovitskiy14nips}
implies an infeasible logit tensor size,
while GAN-based approaches require an additional
generator \cite{salimans16nips,souly17iccv}
or discriminator \cite{hung18bmvc,tsai18cvpr,gao19sensors}.
	Some other approaches require multiple model instances
\cite{rasmus15nips,sajjadi16nips,qiao18eccv,bortsova19miccai}
or accumulated predictions across the dataset \cite{laine17iclr}.
Such designs are less appropriate for dense prediction
since they constrain model capacity.

This paper studies semi-supervised approaches
\cite{sajjadi16nips,zheng16cvpr,laine17iclr,tarvainen17nips,xie20nips}
that require consistent predictions
over input perturbations.
In the considered consistency objective,
input perturbations affect
only one of two model instances,
while the gradient is not propagated
towards the model instance which
operates on the clean (weakly perturbed) input \cite{miyato19pami,xie20nips}.
For brevity,
we refer to the two model instances
as the \textit{perturbed} branch and the \textit{clean} branch.
If the gradient is not computed in a branch,
we refer to it as the \textit{teacher},
and otherwise as the \textit{student}.
Hence, we refer to the considered approach as
one-way consistency with clean teacher.

%
Let $\vec x$ be the input,
$T$ a perturbation to which the ideal model should be invariant,
$h_{\vec\theta}$ the student,
and $h_{\vec\theta'}$ the teacher,
where $\vec\theta'$ denotes
a frozen copy of the student parameters $\vec\theta$.
Then, one-way consistency with clean teacher
can be expressed as a divergence $D$
between the two predictions:
\begin{equation}
	L^\text{ct}_{\vec\theta}
	(\vec x, T) =
	D(h_{\vec{\theta}'}(\vec x),
	h_{\vec{\theta}}(T(\vec x))) \;.
	\label{eq:cons1w}
\end{equation}

\begin{figure}[htb!]
	\centering
	\begin{subfigure}[t]{0.45\textwidth}
		\centering
		\includegraphics[width=1\textwidth]{
			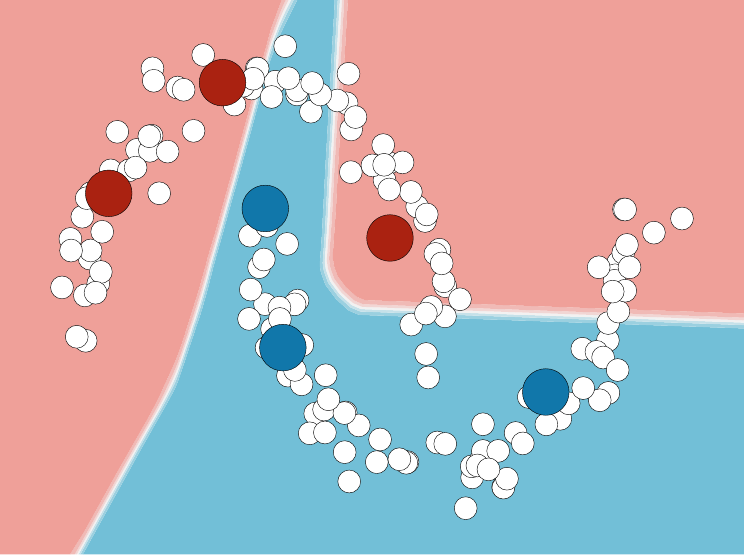}
		\caption{no consistency loss}
		\label{fig:moons-sup}
	\end{subfigure}\;
	\begin{subfigure}[t]{0.45\textwidth}
		\centering
		\includegraphics[width=1\textwidth]{
			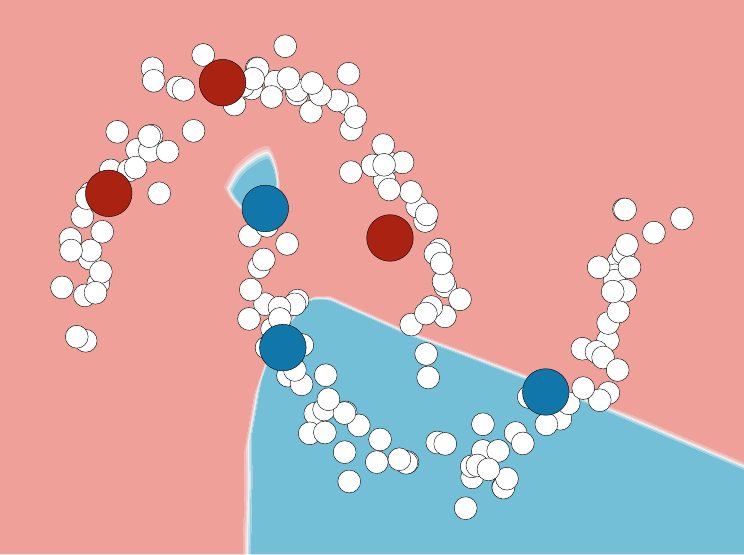}
		\caption{one-way, clean student:
			$D(h_{\vec{\theta}}(\vec x),h_{\vec{\theta}'}(T(\vec x)))$}
		\label{fig:moons-1w-cs}
	\end{subfigure}
	\\\bigskip
	\begin{subfigure}[t]{0.45\textwidth}
		\centering
		\includegraphics[width=1\textwidth]{
			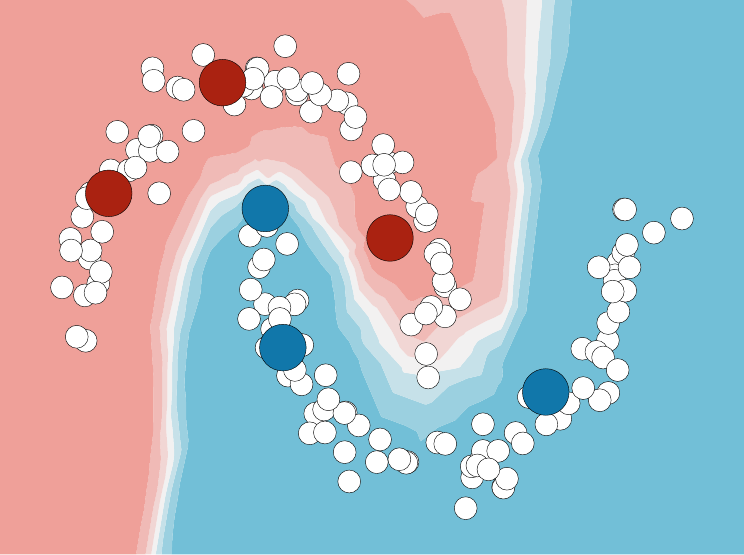}
		\caption{one-way, clean teacher:
			$D(h_{\vec{\theta}'}(\vec x),h_{\vec{\theta}}(T(\vec x)))$}
		\label{fig:moons-1w-ct}
	\end{subfigure}\;
	\begin{subfigure}[t]{0.45\textwidth}
		\centering
		\includegraphics[width=1\columnwidth]{
			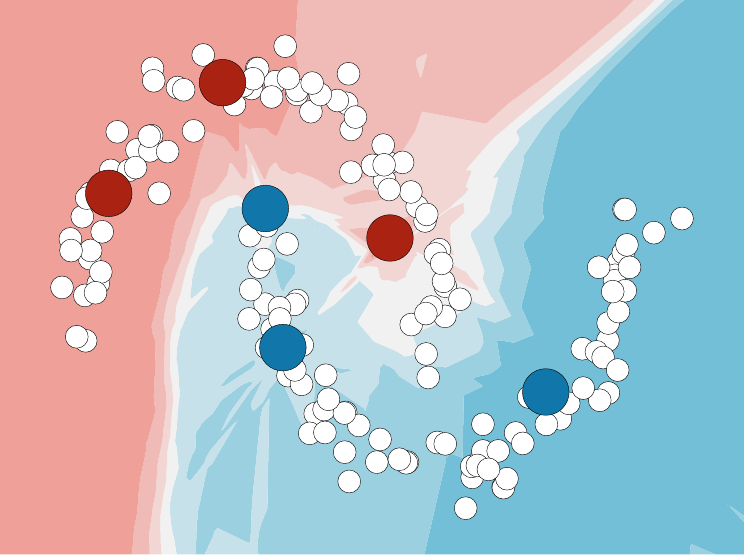}
		\caption{two-way, one input clean:
			$D(h_{\vec{\theta}}(\vec x), h_{\vec{\theta}}(T(\vec x)))$}
		\label{fig:moons-2w-c1}
	\end{subfigure}
	\\\bigskip
	\caption{
		A toy semi-supervised classification problem
		with 6 labeled (red, blue)
		and many unlabeled 2D datapoints (white).
		All setups involve 20\,000 epochs
		of semi-supervised training
		with cross-entropy and default Adam
		optimization hyper-parameters.
		The consistency loss
		was set to none (\protect\subref{fig:moons-sup}),
		one-way with clean student (\protect\subref{fig:moons-1w-cs}),
		one-way with clean teacher (\protect\subref{fig:moons-1w-ct}),
		and two-way with one input clean (\protect\subref{fig:moons-2w-c1}).
		One-way consistency with clean teacher
		outperforms all other formulations.
	}
	\label{fig:moons-cons-variants}
\end{figure}

We argue that the clean teacher approach
is a method of choice
in case of perturbations that are too strong
for standard data augmentation.  
In this setting, perturbed inputs
typically give rise to less reliable predictions
than their clean counterparts.
Figure~\ref{fig:moons-cons-variants} illustrates the advantage
of the clean teacher approach in comparison with
other kinds of consistency on the Two moons dataset.
The clean student experiment (\protect\subref{fig:moons-1w-cs}) shows
that many blue data points
get classified into the red class
due to teacher inputs being pushed towards labeled examples of the opposite class.
This aberration does not occur
when the teacher inputs are clean (\protect\subref{fig:moons-1w-ct}).
Two-way consistency \cite{laine17iclr} (\protect\subref{fig:moons-2w-c1})
can be viewed as a superposition
of the two one-way approaches
and works better than (\protect\subref{fig:moons-1w-cs}),
but worse than (\protect\subref{fig:moons-1w-ct}).
In our experiments, $D$ corresponds to KL divergence.

One-way consistency
is especially advantageous
in the dense prediction context
since it does not require caching latent activations in the teacher.
This allows for better training
in many practical cases where
model capacity is limited by GPU memory \cite{bulo18cvpr,kreso20tits}.
In comparison with two-way consistency
\cite{laine17iclr,bortsova19miccai},
the proposed approach both improves generalization
and approximately halves the training memory footprint.

This paper is an extended version of our
preliminary conference report \cite{grubisic21mva}.
It exposes the elements of our method
in much more detail and complements
them with many new experiments.
In particular, the most important additions are
additional ablation and validation studies,
full-resolution Cityscapes experiments,
and a detailed analysis
of a large-scale experiment that compares
the contribution of coarse labels
with semi-supervised learning on unlabeled images.
The new experiments add more evidence in favor of one-way consistency with respect to other consistency variants, investigate the influence of particular components of our algorithm and various hyper-parameters, and investigate the behavior of the proposed algorithm in different data regimes (higher resolution, additional unlabeled images).

The consolidated paper proposes
a simple and effective method
for semi-supervised semantic segmentation.
One-way consistency with clean teacher
\cite{miyato19pami,xie20nips,french20bmvc}
outperforms the two-way formulation
in our validation experiments.
In addition, it retains the memory footprint
of supervised training
because the teacher activations
depend on parameters that are treated as constants.
Experiments with a standard
convolutional architecture \cite{chen18pami}
reveal that our
photometric and geometric perturbations
lead to competitive generalization performance
and outperform their counterpart
from a recent related work
\cite{french20bmvc}.
A similar advantage can be observed
in experiments with a recent
efficient architecture \cite{orsic19cvpr},
which offers a similar performance 
while requiring an order of magnitude
less computation.
To our knowledge, this is the first account of evaluation of
semi-supervised algorithms for dense prediction
with a model capable of real-time inference.
This contributes to the goals
of Green AI \cite{schwartz20cacm}
by enabling competitive research with less environmental damage.

\section{Related Work} \label{sec:related-work}

Our work spans the fields of dense prediction and semi-supervised learning. The proposed methodology is most related to previous work in semi-supervised semantic segmentation.

\subsection{Dense Prediction}

Image-wide classification models usually achieve efficiency,
spatial invariance, and integration of contextual information
by gradual downsampling of representations
and use of global spatial pooling operations.
However, dense prediction also requires location accuracy. This emphasizes the trade-off between efficiency and quality of high-resolution features in model design.
Some common designs use a classification backbone as a feature encoder and attach a decoder that restores the spatial resolution.
Many approaches seek to enhance contextual information, starting with FCN-8s \cite{long15cvpr}.
UNet \cite{ronneberger15unet} improves spatial details by directly using earlier representations of the encoder in a symmetric decoder.
Further work improves efficiency with lighter decoders \cite{kreso20tits,zhao18eccv}.
Some models use context aggregation modules such as spatial pyramid pooling \cite{zhao17pspnet} and multi-scale inference \cite{zhao18eccv, orsic2021pr}.
DeepLab \cite{chen2018deeplab} increases the receptive field through dilated convolutions and improves spatial details through CRF post-processing.
HRNet \cite{sun2019cvpr} maintains the full resolution throughout the whole model and incrementally introduces parallel lower-resolution branches that exchange information between stages.
Semantic segmentation gains much from ImageNet pre-trained encoders \cite{chen2018deeplab,kreso20tits}.

\subsection{Semi-Supervised Learning}


Semi-supervised methods often rely on some of
the following assumptions about the data distribution \cite{chappelle10mit}: (1)
similar inputs in high density regions correspond to similar outputs (smoothness assumption), (2)
inputs form clusters separated by low-density regions and inputs within clusters are likely to correspond to similar outputs (cluster assumption), and (3) the data lies on a lower-dimensional manifold (manifold assumption).
Semi-supervised methods devise various inductive biases that exploit such regularities for learning from unlabeled data.

Entropy minimization \cite{grandvalet05nips}
encourages high confidence in unlabeled inputs.
Such designs push decision boundaries
towards low-density regions,
under assumptions of clustered data
and prediction smoothness.
Pseudo-label training (or self-training) \cite{yarowsky95hlt,mcclosky06aaai,lee13icmlw}
also encourages high confidence (because of hard pseudo-labels)
as well as consistency with a previously trained teacher.
The basic forms of such algorithms do not achieve
competitive performance on their own \cite{oliver18nips},
but can be effective in conjunction with other approaches \cite{xie20cvpr,miyato19pami}.
Pseudo-labels can be made very effective by confidence-based selection and other processing \cite{yarowsky95hlt,mcclosky06aaai,wang22cvpr}.
Some concurrent work \cite{wang22cvpr}
uses the term pseudo-label
as a synonym for processed teacher prediction
in one-way consistency,
but we do not follow this practice.


Many approaches exploit the smoothness assumption
by enforcing prediction consistency
across different versions of the same input
or different model instances.
Introducing knowledge about equivariance
has been studied for understanding and learning
useful image representations \cite{gerken21arxiv,lenc2018ijcv}
and improving dense prediction \cite{wang20cvpr,cho21cvpr,patel22mia}.
Exemplar training \cite{dosovitskiy14nips}
associates patches with their original images
(each image is a separate surrogate class).
Temporal ensembling \cite{laine17iclr}
enforces per-datapoint consistency
between the current prediction
and a moving average of past predictions.
Mean Teacher \cite{tarvainen17nips} encourages consistency with a teacher
whose parameters are an exponential  moving average
of the student's parameters.

Clusterization of latent representations
can be promoted by penalizing walks which start in a labeled example, pass over an unlabeled example, and end in another example with a different label \cite{hausser17cvpr}.
PiCIE \cite{cho21cvpr} obtains semantically meaningful segmentation without labels
by jointly learning clustering
and representation consistency under photometric and geometric perturbations.

MixMatch \cite{berthelot19nips} encourages consistency
between predictions in different
MixUp perturbations of the same input.
The average prediction is used as a pseudo-label
for all variants of the input.
Deep co-training \cite{qiao18eccv} produces complementary models by encouraging them to be consistent while each is trained on adversarial examples of the other one.

Consistency losses
may encourage trivial solutions,
where all inputs give rise to the same output.
This is not much of a problem in semi-supervised learning
since there the trivial solution is inhibited
through the supervised objective.
Interestingly, recent work 
shows that a variant of simple one-way consistency
evades trivial solutions even
in the context of self-supervised
representation learning \cite{chen2020simsiam,tian21icml} .





Virtual adversarial training (VAT) \cite{miyato19pami}
encourages one-way consistency
between predictions in original datapoints
and their adversarial perturbations.
These perturbations are recovered
by maximizing a quadratic approximation of the prediction divergence
in a small $\mathrm{L}^2$ ball around the input.
Better performance is often obtained by additionally encouraging low-entropy predictions \cite{grandvalet05nips}.
Unsupervised data augmentation (UDA) \cite{xie20nips} also
uses a one-way consistency loss.
FixMatch \cite{kurakin20neurips} shows that pseudo-label selection and processing
can be useful in one-way consistency.
However, instead of adversarial additive perturbations,
they use random augmentations generated by RandAugment.
Different from all previous approaches,
we explore an exhaustive set of consistency formulations.

\subsection{Semi-Supervised Semantic Segmentation}

In the classic semi-supervised GAN (SGAN) setup,
the classifier also acts as a discriminator
which distinguishes between real data
(both labeled and unlabeled)
and fake data produced by the generator \cite{salimans16nips}.
This approach has been adapted for dense prediction
by expressing the discriminator as a segmentation network
that produces dense C+1-way logits \cite{souly17iccv}.
KE-GAN \cite{qi19cvpr} additionally enforces
semantic consistency of neighbouring predictions
by leveraging label-similarity
recovered from a large
text corpus (MIT ConceptNet).
A semantic segmentation model can also
be trained as a GAN generator (AdvSemSeg) \cite{hung18bmvc}.
In this setup,
the discriminator guesses whether
its input is ground truth or generated
by the segmentation network.
The discriminator is also used to choose better
predictions for use as pseudo-labels
for semi-supervised training.
s4GAN + MLMT \cite{mittal19pami}
additionally post-processes the recovered dense predictions
by emphasizing classes identified
by an image-wide classifier trained with Mean Teacher
\cite{tarvainen17nips}.
The authors note that
the image-wide classification
component is not appropriate for datasets such as Cityscapes,
where almost all images contain a large number of classes.

A recent approach
enforces consistency between outputs of redundant decoders
with noisy intermediate representations \cite{ouali20cvpr}.
Other recent work studies pseudo-labeling
in the dense prediction context
\cite{zhu20arxiv,chen20eccv,mendel20eccv}.
Zhu et al. \cite{zhu20arxiv} observe advantages in
hard pseudo-labels.
A recent approach \cite{bortsova19miccai}
proposes a two-way consistency loss,
which is related to the $\upPi$-model
\cite{laine17iclr},
and perturbs both inputs with geometric warps.
However, we show that
perturbing only the student branch
generalizes better
and has a smaller training footprint.
A concurrent work \cite{lai2021cac} successfully applies
a contrastive loss \cite{oord18cpc,he20moco}
between two branches
which receive overlapping crops,
and proposes a pixel-dependent
consistency direction.
Mean Teacher consistency
with CutMix perturbations
achieved state-of-the-art performance
on half-resolution Cityscapes \cite{french20bmvc} prior to this work.
Different than most presented approaches
and similar to \cite{zhu20arxiv,chen20eccv,french20bmvc,yang2021cvpr},  
our method does not increase
the training footprint \cite{bulo18cvpr}.
In comparison with \cite{zhu20arxiv,chen20eccv,yang2021cvpr},
our teacher is updated in each training step,
which eliminates the need
for multiple training episodes.
In comparison with \cite{french20bmvc},
this work proposes a perturbation model
which results in better generalization
and shows that simple one-way consistency
can be competitive with Mean Teacher.
None of the previous approaches
addresses semi-supervised training
of efficient dense prediction models.
We examine the simplest forms of consistency,
explain advantages of perturbing only the student
with respect to other forms of consistency, and
propose a novel perturbation model.
None of the previous approaches considered semi-supervised training of efficient dense-prediction models, nor studied composite perturbations of photometry and geometry.




\section{Method} \label{sec:method}

We formulate dense consistency
as a mean pixel-wise divergence
between corresponding predictions
in the clean image and its perturbed version.
We perturb images with a composition of
photometric and geometric transformations.
Photometric transformations do not disturb
the spatial layout of the input image.
Geometric transformations affect
the spatial layout of the input image,
but the same kind of disturbance
is expected at the model output.
Ideally, our models should exhibit
invariance to photometric transformations
and equivariance to the geometric ones.

\subsection{Notation}

We typeset vectors and arrays in bold, sets in blackboard bold, and we underline random variables.
$\P[\rvar y|\rvec x=\vec x]$ denotes the distribution
of a random variable $\rvar y |\vec x$,
while $\P(y \vert \vec x)$ is a shorthand
for the probability $\P(\rvar y=y \vert \rvec x=\vec x)$.
We denote the expectation of a function of a random variable as e.g. $\E_{\rvec \tau} f(\vec\tau)$. We use similar notation to denote the average over a set: $\E_{\vec x \in \set D} f(\vec x)$.
We denote cross-entropy with $\mathrm{H}_{\rvar y}(\rvar y^*) \coloneqq \E_{y\sim\rvec y^*} \ln \p(\rvar y=y)$, and entropy with  $\mathrm{H}(\rvar y)$ \cite{olah15blog_vit}.
We use Python-like array indexing notation.

We denote the labeled dataset as $\set D_\text{l}$,
and the unlabeled dataset as $\set D_\text{u}$.
We consider input images
$\vec{x}\in [0,1]^{H\times W\times 3}$ and dense labels
$\vec{y}\in \cbr{1\bidot C}^{H\times W}$.
A model instance
maps an image to per-pixel class probabilities:
$h_{\vec{\theta}}(\vec{x})_\ind{i,j,c} = \P(\rvec{y}_\ind{i,j}=c \vert \vec{x}, \vec{\theta})$.
For convenience, we identify output vectors
of class probabilities with distributions:
$h_{\vec{\theta}}(\vec{x})_\ind{i,j} \equiv
\P[\rvec{y}_\ind{i,j} \vert \vec{x}, \vec{\theta}]$.

\subsection{Dense One-Way Consistency}

\begin{figure}[htb!]
	\centering
	\begin{tikzpicture}[scale=2]
		\draw (0,0.5) node[nrect] (x) [] {$\vec x$};
		\draw (1,1) node[nrect] (y) [] {$h_{\vec\theta'}$};
		\draw (2,1) node[nrect] (tp) [] {$T^\text{G}_{\vec\gamma}$};
		\draw (1,0) node[nrect] (xp) [] {$T^\text{G}_{\vec\gamma}\circ T^\text{P}_{\vec\varphi}$};
		\draw (2,0) node[nrect, themeblue, fill=white!95!themeblue] (yp) [] {$h_{\vec\theta}$};
		\draw (3,0.5) node[nrect, themeblue, fill=white!95!themeblue] (D) [] {$D$};

		\node[] (imx) [left=0 of x] {\includegraphics[width=5em]{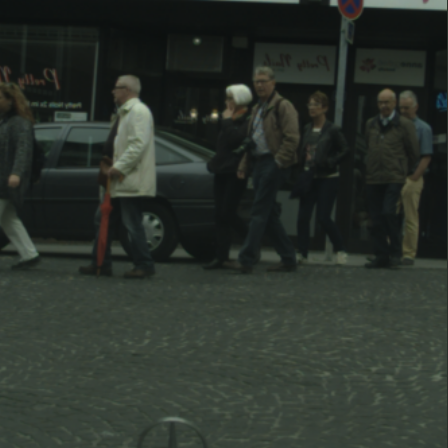}};
		\node (imy) [above=0 of y] {\includegraphics[width=5em]{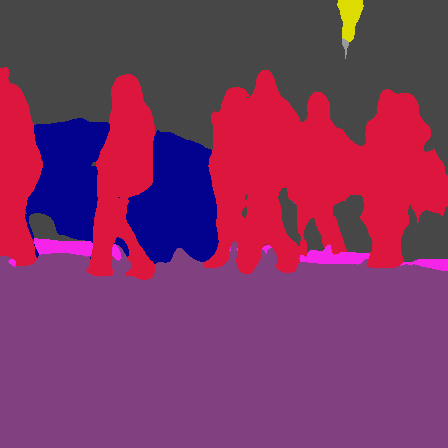}};
		\node (imtp) [above=0 of tp] {\includegraphics[width=5em]{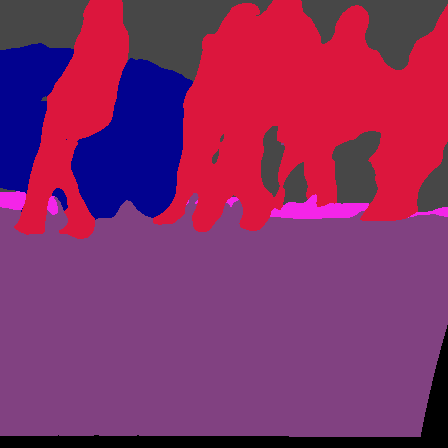}};
		\node (imxp) [below=0 of xp] {\includegraphics[width=5em]{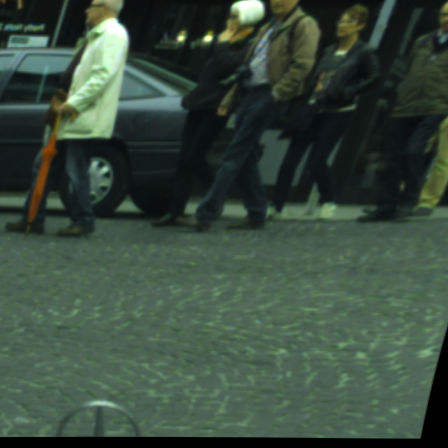}};
		\node (imyp) [below=0 of yp] {\includegraphics[width=5em]{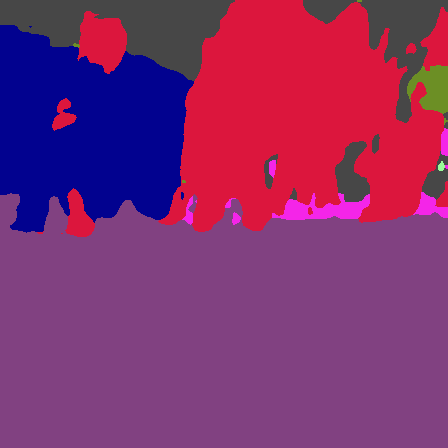}};

		\path (x) edge [dedge] node {} (y);
		\path (y) edge [dedge] node {} (tp);
		\path (x) edge [dedge] node {} (xp);
		\path (xp) edge [dedge] node {} (yp);
		\path (tp) edge [dedge, style={arrows=->||}] node {} (D);
		\path (yp) edge [dedge, themeblue] node {} (D);
	\end{tikzpicture}
	\caption{Dense one-way consistency with clean teacher.
	Top branch: the input is fed to the teacher $h_{\vec\theta'}$. The resulting predictions are perturbed with geometric perturbations $T^\text{G}_{\vec\gamma}$.
	Bottom branch: the input is perturbed with geometric and photometric perturbations and fed to the student $h_{\vec\theta}$. The loss $D$ corresponds to the average pixel-wise KL divergence between the two branches. Gradients are computed only in the blue part of the graph.
	}
	\label{fig:1w-ct-graph}
\end{figure}
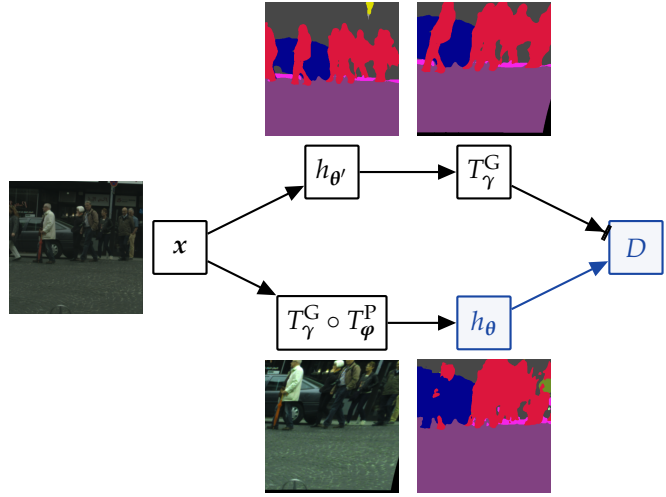

We adapt one-way consistency  
\cite{miyato19pami,xie20nips}
for dense prediction under
our perturbation model
$T_{\vec\tau}=
T^\text{G}_{\vec\gamma} \circ
T^\text{P}_{\vec\varphi}$, where $T^\text{G}_{\vec\gamma}$ is a geometric warp,
$T^\text{P}_{\vec\varphi}$ a per-pixel photometric perturbation,
and $\vec\tau=(\vec\gamma,\vec\varphi)$ perturbation parameters.
$T^\text{G}_{\vec\gamma}$ displaces pixels
with respect to a dense deformation field.
The same geometric warp is applied
to the student input
and the teacher output.
Figure~\ref{fig:1w-ct-graph} illustrates
the computational graph of
the resulting dense consistency loss.
In simple one-way consistency,
the teacher parameters $\vec\theta'$
are a frozen copy of
the student parameter $\vec\theta$.
In Mean Teacher, $\vec\theta'$
is a moving average of $\vec\theta$.
In simple two-way consistency,
both branches use the same $\vec\theta$
and are subject to gradient propagation.

A general semi-supervised training criterion
$L(\vec\theta ; \set D_\text{l}, \set D_\text{u})$
can be expressed as a weighted sum of
a supervised term $L_\text{s}$ over labeled data
and an unsupervised consistency term $L_\text{c}$:
\begin{equation}
	L(\vec\theta ; \set D_\text{l}, \set D_\text{u}) =
	\E_{(\vec x,\vec y)\in \set D_\text{l}}
	L_\text{s}(\vec\theta ; \vec x, \vec y)
	+
	\alpha
	\E_{\vec x\in \set D_\text{u}} \E_{\rvec\tau}
	L_\text{c}(\vec\theta ; \vec{x}, \vec\tau)
	\;.
	\label{eq:semisup-error}
\end{equation}
In our experiments, $L_\text{s}$ is the usual
mean per-pixel cross entropy
with $\mathrm{L}^2$ regularization.
We stochastically estimate
the expectation over perturbation parameters $\rvec\tau$
with one sample per training step.

We formulate the unsupervised term $L_\text{c}$
at pixel $(i,j)$
as a one-way divergence $D$ between
the prediction in the perturbed image
and its interpolated correspondence
in the clean image.
The proposed loss encourages the trained model
to be equivariant to $T^\text{G}_{\vec\gamma}$ and
invariant to $T^\text{P}_{\vec\varphi}$:
\begin{align}
	L_{\text c}^{i,j}(\vec\theta; \vec x, \vec\tau) =
	D(
	T^\text{G}_{\vec{\gamma}}(
	h_{\vec\theta'}(\vec x))_\ind{i,j},
	h_{\vec\theta}(
	(T^\text{G}_{\vec\gamma} \circ T^\text{P}_{\vec\varphi})
	(\vec x))_\ind{i,j}
	)
	\;.
	\label{eq:1w-ct-objective}
\end{align}
We use a validity mask $\vec v^{\vec\gamma}\in\cbr{0,1}^{H\times W}$,
$\vec v^{\vec\gamma}_\ind{i,j} =
\enbbracket{
	T^\text{G}_{\vec{\gamma}}(\cvec 1_{H\times W})_\ind{i,j} = 1
}$ to ensure that the loss is unaffected by padding sampled from outside of $\sbr{1,H}\times\sbr{1,W}$.
A vector produced by
$T^\text{G}_{\vec{\gamma}}
(h_{\vec\theta}(\vec x))_\ind{i,j}$
represents a valid distribution
wherever $\vec v^{\vec\gamma}_\ind{i,j}=1$.
Finally, we express the consistency term $L_\text{c}$
as a mean contribution over all pixels:
\begin{equation}
	L_\text{c}(\vec\theta; \vec x, \vec\tau)
	=
	\frac{1}{\sum(\vec v^{\vec\gamma})}
	\sum_{i,j} \vec v^{\vec\gamma}_\ind{i,j}
	L_{\text{c}}^{i, j}
	(\vec\theta ; \vec x, \vec\tau)
	\;
	.
	\label{eq:cons-loss}
\end{equation}

\begin{figure}[htb!]
	\centering
	\includegraphics[width=0.8\textwidth]{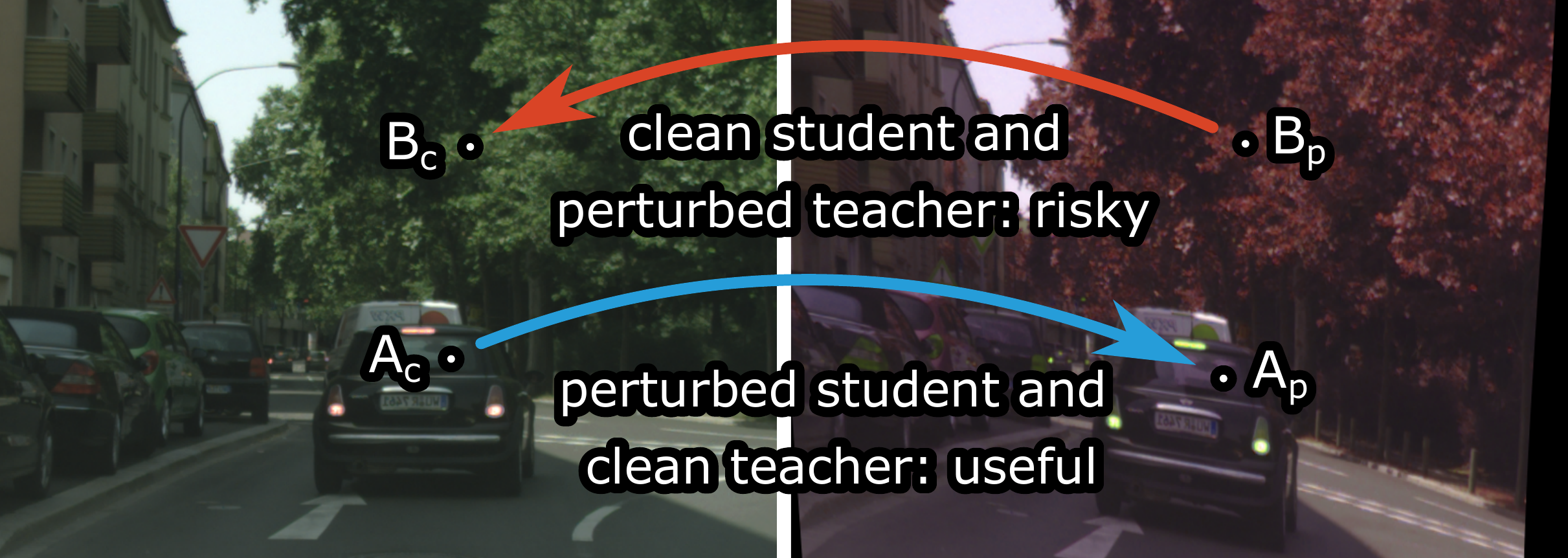}
	\caption{
		Two variants of one-way consistency training
		on a clean image (left)
		and its perturbed version (right).
		The arrows designate information flow from the teacher to the student.
		The proposed clean-teacher formulation
		trains in the perturbed pixels ($\mathrm{A_p}$)
		according to the corresponding predictions
		in the clean image ($\mathrm{A_c}$).
		The reverse formulation
		(training in $\mathrm{B_c}$ according to the prediction in $\mathrm{B_p}$)
		worsens performance
		since strongly perturbed images
		often give rise to less accurate predictions.
	}
	\label{fig:1w_motivation}
\end{figure}

Recall that the gradient is not computed
with respect to $\vec\theta'$.
Consequently, $L_\text{c}$
allows gradient propagation
only towards the perturbed image.
We refer to such training
as one-way consistency with clean teacher (and perturbed student).
Such formulation provides two distinct advantages
over other kinds of consistency.
First, predictions in perturbed images are
pulled towards predictions in clean images.
This improves generalization
when the perturbations are stronger
than data augmentations used in $L_\text{s}$
(cf. Figures~\ref{fig:1w_motivation} and \ref{fig:moons-cons-variants}).
Second, we do not have to cache
teacher activations during training
since the gradients propagate
only towards the student branch.
Hence, the proposed semi-supervised objective
does not constrain model complexity
with respect to the supervised baseline.


We use KL divergence as a principled choice
for $\mathit{D}$:
\begin{equation}
	\mathit{D}(\rvar y, \tilde{\rvar y})
	\coloneqq \E_{\rvar y}
	\ln\frac{\P(\rvar y= y)}
	{\P(\tilde{\rvar y}=y)}
	= \mathrm{H}_{\tilde{\rvar y}}(\rvar y) - \mathrm{H}(\rvar y) \;.
	\label{eq:kldiv}
\end{equation}
Note that the entropy term $-\mathrm{H}(\rvar y)$ 
does not affect parameter updates
since the gradients are not propagated through $\vec\theta'$.
Hence, one-way consistency does not encourage
increasing entropy of model predictions in clean images.
Several researchers have observed improvement
after adding an  entropy minimization term \cite{grandvalet05nips}
to the consistency loss \cite{miyato19pami,xie20nips}.
This practice did not prove beneficial in our initial experiments.

Note that two-way consistency \cite{laine17iclr,bortsova19miccai}
would be obtained by replacing
$\vec\theta'$ with $\vec\theta$.
It would require caching latent activations
for both model instances,
which approximately doubles the training footprint
with respect to the supervised baseline.
This would be undesirable due to constraining the
feasible capacity of the deployed models
\cite{bulo18cvpr,huang22pami}.

We argue that consistency with clean teacher
generalizes better than consistency with clean student
since strong perturbations
may push inputs
beyond the natural manifold and
spoil predictions (cf. Figure~\ref{fig:moons-cons-variants}).
Moreover, perturbing both branches sometimes results
in learning to map all perturbed pixels to similar
arbitrary predictions (e.g. always the same class) \cite{grill2020byol}.
Figure~\ref{fig:1w_motivation} illustrates that
consistency training has the best chance to succeed
if the teacher is applied to the clean image,
and the student learns on the perturbed image.

\subsection{Photometric Component of the Proposed Perturbation Model}
\label{subsec:photometric}

We express pixel-level photometric transformations
as a composition of five simple perturbations
with five image-wide parameters
$\vec\varphi = (b,s,h,c,\vec\pi)$.
These perturbations
are applied in each pixel
in the following order:
(1) brightness is shifted by adding $b$ to all channels,
(2) saturation is multiplied with $s$,
(3) hue is shifted by addition with $h$,
(4) contrast is modulated
by multiplying all channels with $c$, and
(5) RGB channels are randomly permuted
according to $\vec\pi$.
The resulting compound transformation
$T^\text{P}_{\vec\varphi}$
is independently applied
to all image pixels.

Our training procedure randomly picks image-wide parameters $\vec\varphi$
for each unlabeled image.
The parameters are sampled as follows:
$b\sim\mathcal U(-0.25, 0.25)$,
$s\sim\mathcal U(0.25, 2)$,
$h\sim\mathcal U(-36^\circ, 36^\circ)$,
$c\sim\mathcal U(0.25, 2)$, and $\vec\pi\sim \mathcal U(\cset S_3)$,
where $\cset S_3$ represents the set of all $6$ 3-element permutations.

\subsection{Geometric Component of the Proposed Perturbation Model}
\label{subsec:tps}

We formulate a fairly general class
of parametric geometric transformations
by leveraging thin plate splines (TPS) \cite{duchon77ctfsv,bookstein89pami}.
We consider the 2D TPS warp $f\colon \cset R^2\to\cset R^2$
which maps each image coordinate pair $\vec q$
to the relative 2D displacement
of its correspondence $\vec q'$:
	$f(\vec q) = \vec q' - \vec q$. 
TPS warps minimize the bending energy (curvature)
$\int_{\dom(f)} \norm{\pd[2]{f(\vec q)}{\vec q}}_\text{F}^2 \dif \vec q$
given a set of control points
and their displacements
$\cbr{(\vec c_i, \vec d_i)\colon i=1\bidot n} \subset \cset R^2\times\cset R^2$.
In simple words, a TPS warp produces
a smooth deformation field which
optimally satisfies all constraints
$f(\vec c_i) = \vec d_i$.
In the 2D case, the solution of the TPS problem takes the following form:
\begin{equation} \label{eq:tps_2d}
	f(\vec q)
	=
	\vec A
	\begin{bmatrix} 1 \\ \vec q \end{bmatrix}
	+
	\vec W
	\begin{bmatrix}\phi(\norm{\vec q - \vec c_i})\end{bmatrix}_{i=1\bidot n}^\tp \;.
\end{equation}
In the above equation, $\vec q$ denotes
a 2D coordinate vector to be transformed,
$\vec A$ is a $2\times 3$ affine transformation matrix,
$\vec W$ is a $2\times n$
control point coefficient matrix,
and $\phi(r)=r^2\ln(r)$.
Such a 2D TPS warp is equivariant to rotation and translation \cite{bookstein89pami}.
That is, $f(T(\vec q)) = T(f(\vec q))$
for every composition of
rotation and translation $T$.

TPS parameters $\vec A$ and $\vec W$ can be determined
as a solution of a standard linear system
which enforces deformation constraints
$(\vec c_i, \vec d_i)$,
and square-integrability of second derivatives of $f$.
When we determine $\vec A$ and $\vec W$,
we can easily transform entire images.

We first consider images
as continuous domain functions
and later return to images as arrays from $[0,1]^{H\times W\times 3}$.
Let $I \colon \dom(I)\to\intcc{0,1}^3$ be the original image of size $(W, H)$, where $\dom(I)=\intcc{0,W}\times\intcc{0,H}$.
Then the transformed image $I'$ can be expressed as
\begin{equation}
	I'(\vec q+f(\vec q)) = \begin{cases} I(\vec q), & \vec q \in \dom(I), \\ \cvec{0}, & \text{otherwise.} \end{cases}
\end{equation}

The resulting formulation is known as
forward warping \cite{szeliski2010computer}
and is tricky to implement.
We, therefore, prefer to recover
the reverse transformation $\tilde f$,
which can be done
by replacing each control point $\vec c_i$
with $\vec c_i'=\vec c_i+\vec d_i$.
Then the transformed image is:
\begin{equation}
	\tilde{I}(\vec q') =
	\begin{cases}
		I(\vec q' - \tilde f(\vec q')),
		& \vec q' - \tilde f(\vec q') \in \dom(I),
		\\
		\cvec{0}, & \text{otherwise.}
	\end{cases}
\end{equation}
This formulation is known as
backward warping \cite{szeliski2010computer}.
It can be easily implemented
for discrete images
by leveraging bilinear interpolation.
Contemporary frameworks already include
the implementations for GPU hardware.
Hence, the main difficulty is
to determine the TPS parameters
by solving two linear systems
with $(n+3)\times (n+3)$ variables \cite{bookstein89pami}.

In our experiments, we use $n=4$ control points
corresponding to the centers of the four image quadrants:
$\del{\vec c_1', .., \vec c_4'} =
\del{
	\sbr{\frac{1}{4}H, \frac{1}{4}W}^\tp,
	..,
	\sbr{\frac{3}{4}H,\frac{3}{4}W}^\tp}$.
The parameters of our geometric transformation
are four 2D displacements
$\vec\gamma = \del{\vec d_1, .., \vec d_4}$.
Let $f_{\vec\gamma}$ denote the resulting TPS warp.
Then we can express our transformation as
$T^\text{G}_{\vec\gamma}(\vec x) = \mathrm{backward\_warp}(
\vec x, f_{\vec\gamma})$. 

Our training procedure picks a random $\vec\gamma$
for each unlabeled image.
Each displacement is sampled from
a 2D normal distribution
$\mathcal N\del{\cvec 0_2, 0.05 H\cdot \cvec I_2}$,
where $H$ is the training crop height.

\subsection{Training Procedure}

Algorithm~\ref{alg:training} sketches a procedure for recovering
gradients of the proposed semi-supervised loss~\eqref{eq:semisup-error}
on a mixed batch of labeled
and unlabeled examples.
For simplicity, we make the following changes in notation here:
$\vec x_\text{l}$ and $\vec y_\text{l}$ are batches of size $B_\text{l}$, $\vec x_\text{u}$, $\vec\gamma$ and $\vec\varphi$ batches of size $B_\text{u}$, and all functions are applied to batches.
The algorithm computes the gradient of the supervised loss,
discards cached activations,
computes the teacher predictions,
applies the consistency loss~\eqref{eq:1w-ct-objective}, and
finally accumulates the gradient contributions of the two losses.
Backpropagation through one-way consistency with clean teacher
requires roughly as much caching
as the supervised baseline.
Hence, our approach constrains
the model complexity
much less than
the two-way consistency.

\begin{algorithm}[%
	caption={Evaluation of the gradient of the proposed semi-supervised loss given perturbation parameters ($\vec\gamma$, $\vec\varphi$) on a mixed batch of labeled ($\vec x_\text{l}$, $\vec y_\text{l}$) and unlabeled ($\vec x_\text{u}$) examples. $\mathrm{CE}$ denotes mean cross entropy, while $\mathrm{KL\_masked}$ denotes mean KL divergence over valid pixels.},
	%float, floatplacement=hbt!,
	label={alg:training}]
# $\vec x_\text{u}$, $\vec\gamma$, $\vec\varphi$, $\vec p_\text{t}$, $\vec p_\text{s}$, and $\vec v$ contain batches of $B_\text{u}$ elements.
# $\vec x_\text{l}$, $\vec y$, and $\vec p_\text{l}$ contain batches of $B_\text{l}$ elements.
procedure compute_loss_gradient($h, \vec\theta, \vec x_\text{l}, \vec y_\text{l}, \vec x_\text{u}, \vec\gamma, \vec\varphi$):
   $\vec\theta' \gets \mathrm{frozen\_copy}(\vec\theta)$  # simple one-way

   # supervised loss
   $\vec p_\text{l} \gets h_{\vec\theta}(\vec x_\text{l})$
   $L_\text{s} \gets \mathrm{CE}(\vec y_\text{l}, \vec p_\text{l})$
   $\vec g \gets \nabla_{\vec\theta}L_\text{s}$  # clears cached activations

   # unsupervised loss
   with no_grad():  # activations not cached here
      $\vec p_\text{t} \gets T^\text{G}_{\vec\gamma}(h_{\vec\theta'}(\vec x_\text{u}))$  # clean teacher
   $\vec p_\text{s} \gets h_{\vec\theta}((T^\text{G}_{\vec\gamma} \circ T^\text{P}_{\vec\varphi})(\vec x_\text{u}))$  # perturbed student
   $\vec v \gets \lfloor T^\text{G}_{\vec\gamma}(\cvec 1_{B_\text{u}\times H\times W}) \rfloor$  # validity mask
   $L_\text{c} \gets \alpha\cdot \mathrm{KL\_masked}(\vec p_\text{t}, \vec p_\text{s}, \vec v)$
   $\vec g \gets \vec g + \nabla_{\vec\theta}L_\text{c}$

   return $\vec g$
\end{algorithm}

\begin{figure}[htb!]
	\centering
	\includegraphics[scale=1]{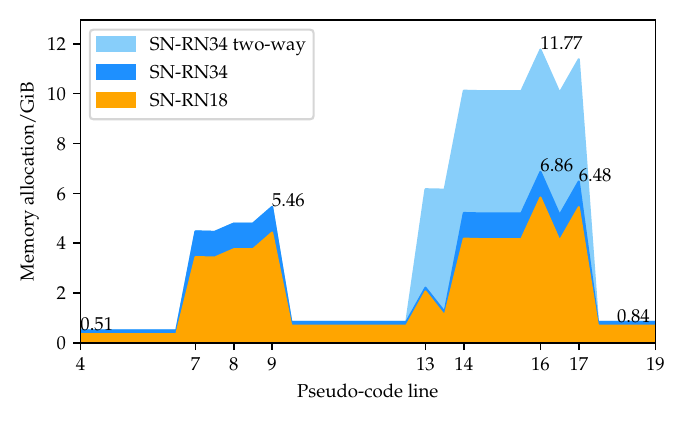}
	\caption{
    Total GPU memory allocation
	during and after execution of particular lines
	from Algorithm~\ref{alg:training}
	during the 2nd iteration of training.
	Our PyTorch implementations involve
	SwiftNet-RN18 and SwiftNet-RN34 models
	with one-way and two-way consistency,
	$768\times768$ crops, and batch sizes
	$(B_\text{l}, B_\text{u}) = (8,8)$.
	Line 9 computes the supervised gradient.
	Line 13 computes the teacher output
	(without caching interemediate activations).
	Lines 16 and 17 compute
	the consistency loss
	and its gradient.
	}
	\label{fig:cs_memory_alloc}
\end{figure}

Figure~\ref{fig:cs_memory_alloc} illustrates
GPU memory allocation
during a semi-supervised training iteration
of a SwiftNet-RN34 model
with one-way and two-way consistency.
We recovered these measurements
by leveraging the following functions
of the \texttt{torch.cuda} package:
\texttt{max\_memory\_allocated},
\texttt{memory\_allocated},
\texttt{reset\_peak\_memory\_stats}, and
\texttt{empty\_cache}.
The training was carried out
on a RTX A4500 GPU.
Numbers on the x-axis correspond to lines
of the pseudo-code in Algorithm~\ref{alg:training}.
Line 9 backpropagates through the supervised loss
and caches the gradients.
The memory footprint briefly peaks
due to temporary storage
and immediately declines
since PyTorch automatically releases
all cached activations
immediately after the backpropagation.
Line 13 computes the teacher output.
This step does not
cache intermediate activations
since we freeze parameters
with \texttt{torch.no\_grad}.
Line 16 computes the unsupervised loss,
which requires caching of activations
on a large spatial resolution.
The memory footprint briefly peaks
since we delete perturbed inputs
and teacher predictions
immediately after line 16
(for simplicity, we omit
opportunistic deletions
from Algorithm 1).
Line 17 triggers the backpropagation algorithm
and accumulates the gradients
of the consistency loss.
The memory footprint briefly peaks
due to temporary storage
and immediately declines
due to automatic deletion
of cached activations.
At this point, the memory footprint
is slightly greater than at line 4
since we still hold the supervised predictions
in order to accumulate recognition performance
on the training dataset.

The ratio between memory allocations
at lines 16 and 9 reveals the
relative memory overhead
of our semi-supervised approach.
Note that the absolute overhead
is model independent since it corresponds
to the total size of perturbed inputs and predictions,
and intermediate results of dense KL-divergence.
On the other hand,
the memory footprint of the supervised baseline
is model dependent since it reflects
computational complexity of the backbone.
Consequently, the relative overhead
approaches 1 as the model size increases,
and is around 1.26 for SwiftNet-RN34.

\section{Results}\label{sec:results} \label{sec:experiments}

Our experiments evaluate one-way consistency with clean teacher
and a composition of photometric and geometric perturbations
($T_\gamma^G \circ T_\phi^P$). We compare our approach
with other kinds of consistency
and the state of the art in semi-supervised
semantic segmentation.
We denote simple one-way consistency as "simple",
Mean Teacher \cite{tarvainen17nips} as "MT",
and our perturbations as "PhTPS".
In experiments that compare consistency variants,
"1w" denotes one-way, "2w" denotes two way, "ct" denotes clean teacher,
"cs" denotes clean student, and "2p" denotes both inputs perturbed.
We present semi-supervised experiments
in several semantic segmentation setups
as well as in image-classification setups on CIFAR-10.
Our implementations are based on the PyTorch framework \cite{pytorch19}.



\subsection{Experimental Setup}\label{sec:experimental-setup}

\textit{Datasets.} We perform semantic segmentation on
Cityscapes \cite{cordts15cvprw}, and image classification on CIFAR-10.
Cityscapes contains
2975 training, 500 validation and
1525 testing images
with resolution $1024\times2048$.
Images are acquired from a moving vehicle
during daytime and fine weather conditions.
We present half-resolution and
full-resolution experiments.
We use bilinear interpolation for images
and nearest neighbour subsampling for labels.
Some experiments on Cityscapes also use
the coarsely labeled Cityscapes subset ("train-extra")
that contains 19998 images.
CIFAR-10 consists of 50000 training
and 10000 test images
of resolution $32\times32$.

\textit{Common setup.} We include both unlabeled and labeled images in $\set D_\text{u}$,
which we use for the consistency loss.
We train on batches of $B_\text{l}$ labeled
and $B_\text{u}$ unlabeled images.
We perform
$\left\lfloor
\envert{\set D_\text{l}}/B_\text{l}
\right\rfloor$
training steps per epoch.
We use the same perturbation model
across all datasets and tasks
(TPS displacements are proportional to image size),
which is likely suboptimal \cite{cubuk20nips}.
Batch normalization statistics are updated only
in non-teacher model instances with clean inputs
except for full-resolution Cityscapes,
for which updating the statistics in the perturbed student performed better in our validation experiments (cf. Appendix~\ref{sec:appendix-additional-experiments}).
The teacher always uses the estimated population statistics,
and does not update them.
In Mean Teacher, the teacher uses an exponential moving average
of the student's estimated population statistics.

\textit{Semantic segmentation setup.} Cityscapes experiments involve the following models:
SwiftNet with ResNet-18 (SwiftNet-RN18) or ResNet-34 (SwiftNet-RN34), and
DeepLab v2 with a ResNet-101 backbone.
We initialize the backbones with
ImageNet pre-trained parameters.
We apply random scaling, cropping, and horizontal flipping
to all inputs and segmentation labels. We refer to such examples as \textit{clean}.
We schedule the learning rate according to
$e\mapsto\eta \cos(e\pi/2)$,
where $e\in\sbr{0..1}$ is the fraction of epochs completed.
This alleviates the generalization drop
at the end of training with
standard cosine annealing \cite{loschilov16sgdr}.
We use learning rates $\eta=4\cdot10^{-4}$ for randomly initialized parameters
and $\eta=10^{-4}$ for pre-trained parameters.
We use Adam with $(\beta_1,\beta_2)=(0.9, 0.99)$.
The L$^2$ regularization weight in supervised experiments is
$10^{-4}$ for randomly initialized and
$2.5\cdot10^{-5}$ for pre-trained parameters \cite{orsic19cvpr}.
We have found that such L$^2$ regularization is too strong
for our full-resolution semi-supervised experiments.
Thus, we use a 4$\times$ smaller weight there.
Based on early validation experiments,
we use $\alpha=0.5$ unless stated otherwise.
Batch sizes are $(B_\text{l}, B_\text{u})=(8,8)$
for SwiftNet-RN18 \cite{orsic19cvpr}
and $(B_\text{l}, B_\text{u})=(4,4)$
for DeepLab v2 (ResNet-101 backbone) \cite{chen18pami}.
The batch size in corresponding supervised experiments
is $B_\text{l}$.

In half-resolution Cityscapes experiments
the size of crops is $448\times 448$
and the logarithm of the scaling factor is sampled from $\mathcal{U}(\ln(1.5^{-1}),\ln(1.5))$.
We train SwiftNet for $200\cdot\frac{2975}{\envert{\set D_\text{l}}}$ epochs ($200$ epochs or $74200$ iterations when all labels are used),
and DeepLab v2 for $100\cdot\frac{2975}{\envert{\set D_\text{l}}}$ epochs ($100$ epochs or $74300$ iterations when all labels are used).
In comparison with SwiftNet-RN18,
DeepLab v2 incurs a 12-fold per-image slowdown
during supervised training.
However, it also requires less epochs
since it has very few parameters
with random initialization.
Hence, semi-supervised DeepLab v2
trains more than 4$\times$ slower than SwiftNet-RN18 on RTX 2080Ti.
Appendix~\ref{subsec:performance-characteristics}
presents more detailed comparisons
of memory and time requirements
of different semi-supervised algorithms.

Our full-resolution experiments
only use SwiftNet models.
The crop size is $768\times 768$
and the spatial scaling is sampled from $\mathcal{U}(2^{-1},2)$.
The number of epochs is 250
when all labels are used.
The batch size is $8$ in supervised experiments,
and $(B_\text{l}, B_\text{u})=(8,8)$
in semi-supervised experiments.

Appendix~\ref{subsec:hyper-parameter-overview} presents an overview and comparison of hyper-parameters with other consistency-based methods that are compared in the experiments.

\textit{Classification setup.} Classification experiments target CIFAR-10
and involve the Wide ResNet model WRN-28-2
with standard hyper-parameters \cite{zagoruyko16bmvc}.
We augment all training images
with random flips, padding and random cropping. 
We use all training images (including labeled images) in $\set D_\text{u}$ for the consistency loss.
Batch sizes are $(B_\text{l}, B_\text{u})=(128,640)$.
Thus, the number of iterations per epoch is
$\left\lfloor\frac{\envert{\set D_\text{l}}}{128}\right\rfloor$.
For example, only one iteration is performed
if $\envert{\set D_\text{l}}=250$.
We run $1000\cdot \frac{4000}{\envert{\set D_\text{l}}}$
epochs in semi-supervised, and $100$ epochs in supervised training.
We use default VAT hyper-parameters $\xi=10^{-6}$, $\epsilon=10$, $\alpha=1$ \cite{miyato19pami}.
We perform photometric perturbations as described, and sample TPS displacements from $\mathcal N(\cvec 0, 3.2\cdot\cvec I_2)$.

\textit{Evaluation.} We report generalization performance
at the end of training.
We report sample means and sample standard deviations (with Bessel's correction) of the corresponding evaluation metric ($\mathrm{mIoU}$ or classification accuracy) of $5$ training runs, evaluated on the corresponding validation dataset.

\subsection{Semantic Segmentation on Half-Resolution Cityscapes}

Table~\ref{tab:cs_half} compares our approach
with the previous state of the art.
We train using different proportions of training labels and evaluate mIoU on half-resolution Cityscapes val.
The top section presents the previous work
\cite{mittal19pami,hung18bmvc,mendel20eccv,french20bmvc}.
The middle section presents our experiments based on DeepLab v2 \cite{chen18pami}.
Note that here we outperform some previous work
due to more involved training (as described in Section~\ref{sec:experimental-setup}).
since that would be a method of choice
in all practical applications.
Hence, we get consistently
greater performance.
We perform a proper comparison with \cite{french20bmvc} by
using our training setup in combination with their method.
Our MT-PhTPS outperforms MT-CutMix
with L2 loss and confidence thresholding
when 1/4 or more labels are available,
while underperforming with 1/8 labels.

The bottom section involves the efficient model SwiftNet-RN18.
Our perturbation model outperforms CutMix
both with simple consistency, as well as with Mean Teacher.
Overall, Mean Teacher outperforms simple consistency.
We observe that DeepLab v2
and SwiftNet-RN18
get very similar benefits from the consistency loss.
SwiftNet-RN18 comes out
as a method of choice 
due to about 12$\times$ faster inference
than DeepLab v2 with ResNet-101 on RTX 2080Ti (see Appendix~\ref{subsec:performance-characteristics} for more details).
Experiments from the middle and the bottom section
use the same splits
to ensure a fair comparison.

\begin{table}[htb!]
	\caption{
		Semantic segmentation performance (mIoU/\%)
		on half-resolution Cityscapes val
		after training with different proportions of labeled data.
		The top section reviews experiments
		from previous work.
		The middle section presents our experiments with DeepLab v2.
		%
		The bottom section presents
		our experiments
		with SwiftNet-RN18.
		We run experiments
		across 5 different dataset splits
		and report mean mIoUs with standard deviations.
		The subscript "$\sim$\cite{french20bmvc}"
		denotes training with $\mathrm{L^2}$ loss,
		confidence thresholding, and $\alpha=1$,
		as proposed in \cite{french20bmvc}.
		Best results overall are bold,
		and best results within sections are underlined.
	}
	\centering
	\begin{tabular}{@{\;}l@{\;\;}c@{\;\;}
			c@{\;\;}c@{\;\;}c@{\;\;}c@{\;}}
		\toprule
		\multirow{2}{*}{\textbf{Method}} & \multicolumn{4}{c}{\textbf{Label proportion}}
		\\
		& $1/8$  &  $1/4$  &  $1/2$  & $1/1$ \\
		\midrule
		DLv2-RN101 supervised \cite{mittal19pami,french20bmvc} &
		$56.2_{\phantom{0.0}}$ & $60.2_{\phantom{0.0}}$ & $64.6^1_{\phantom{0.0}}$ & $66.0_{\phantom{0.0}}$
		\\
		DLv2-RN101 s4GAN+MLMT \cite{mittal19pami} & 
		$59.3_{\phantom{0.0}}$ & $61.9_{\phantom{0.0}}$ & -- & $65.8_{\phantom{0.0}}$
		\\
		DLv2-RN101 supervised \cite{hung18bmvc} &
		$55.5_{\phantom{0.0}}$  & $59.9_{\phantom{0.0}}$  & $64.1_{\phantom{0.0}}$ & $66.4_{\phantom{0.0}}$
		\\
		DLv2-RN101 AdvSemSeg \cite{hung18bmvc} &
		$58.8_{\phantom{0.0}}$  & $62.3_{\phantom{0.0}}$  & $65.7_{\phantom{0.0}}$ & $67.7_{\phantom{0.0}}$
		\\
		DLv2-RN101 supervised \cite{mendel20eccv} &
		$56.0_{\phantom{0.0}}$ & $60.5_{\phantom{0.0}}$ & -- & $66.0_{\phantom{0.0}}$
		\\
		DLv2-RN101 ECS \cite{mendel20eccv} &
		$\underline{60.3}_{\phantom{0.0}}$ & $\underline{63.8}_{\phantom{0.0}}$ & -- & $\underline{67.7}_{\phantom{0.0}}$
		\\
		DLv2-RN101 MT-CutMix \cite{french20bmvc} &
		$\underline{60.3}_{1.2}$ & $\underline{63.9}_{0.7}$ & -- & $\underline{67.7}_{0.4}$
		\\
		\midrule
		DLv2-RN101 supervised &
		$56.4_{0.4}$ & $61.9_{1.1}$ & $66.6_{0.6}$ & $69.8_{0.4}$
		\\
		DLv2-RN101 MT-CutMix\textsubscript{$\sim$\cite{french20bmvc}} &
		$\mathbf{63.2}_{1.4}$  & $65.6_{0.8}$  & $67.6_{0.4}$ & $70.0_{0.3}$
		\\
		DLv2-RN101 MT-PhTPS &
		$61.5_{1.0}$  & $\mathbf{66.4}_{1.1}$ & $\mathbf{69.0}_{0.6}$ & $\mathbf{71.0}_{0.7}$
		\\
		\midrule
		SN-RN18 supervised &
		$55.5_{0.9}$ & $61.5_{0.5}$ & $66.9_{0.7}$ & $70.5_{0.6}$
		\\
		SN-RN18 simple-CutMix &
		$59.8_{0.5}$ & $63.8_{1.2}$ & $67.0_{1.4}$ & $69.3_{1.1}$
		\\
		SN-RN18 simple-PhTPS &
		$60.8_{1.6}$ & $64.8_{1.5}$ & $\mathbf{68.8}_{0.7}$ & $\mathbf{71.1}_{0.9}$
		\\
		SN-RN18 MT-CutMix\textsubscript{$\sim$\cite{french20bmvc}} &
		$61.6_{0.9}$  & $64.6_{0.5}$ & $67.6_{0.7}$ & $69.9_{0.6}$
		\\
		SN-RN18 MT-CutMix &
		$59.3_{1.3}$  & $63.3_{1.0}$ & $66.8_{0.6}$ & $69.7_{0.5}$
		\\
		SN-RN18 MT-PhTPS &
		$\underline{62.0}_{1.3}$ & $\mathbf{66.0}_{1.0}$ & $\mathbf{69.1}_{0.5}$ & $\mathbf{71.2}_{0.7}$
		\\
		\bottomrule
	\end{tabular}
	\label{tab:cs_half}
\end{table}

Now we present
ablation and hyper-parameter validation studies
for simple-PhTPS consistency with SwiftNet-RN18.
Table~\ref{tab:cs_half_ablation} presents ablations
of the perturbation model,
and also includes supervised training with
PhTPS augmentations
in one half of each mini-batch
in addition to standard jittering.
Perturbing the whole mini-batch
with PhTPS in supervised training
did not improve upon the baseline.
We observe that perturbing half of each mini-batch
with PhTPS in addition to standard jittering
improves supervised performance,
but quite less than semi-supervised training.
Semi-supervised experiments suggest that
photometric perturbations (Ph) contribute most,
and that geometric perturbations (TPS) are not useful
when there is $1/2$ or more of the labels.

\begin{table}[htb!]
	\caption{
		Ablation experiments
		on half-resolution Cityscapes val (mIoU/\%)
		with SwiftNet-RN18.
		Subscripts denote the difference from the supervised baseline.
		The label "supervised PhTPS-aug" denotes
		supervised training where
		half of each mini-batch
		is perturbed with PhTPS.
		The bottom three rows compare PhTPS with
		Ph (only photometric) and TPS (only geometric)
		under simple one-way consistency.
		We present means of experiments
		on 5 different dataset splits.
		Numerical subscripts are differences
		with respect to the supervised baseline.
	}
	\centering
	\begin{tabular}{@{\;}l@{\;\;}c@{\;\;}
			c@{\;\;}c@{\;\;}c@{\;\;}c@{\;}}
		\toprule
		\multirow{2}{*}{\textbf{Method}} & \multicolumn{4}{c}{\textbf{Label proportion}}
		\\
		& $1/8$ & $1/4$ & $1/2$ & $1/1$ \\
		\midrule
		SN-RN18 supervised &
		$55.5_{\phantom{+0.0}}$ & $61.5_{\phantom{+0.0}}$ & $66.9_{\phantom{+0.0}}$ & $70.5_{\phantom{+0.0}}$
		\\
		SN-RN18 supervised PhTPS-aug &
		$56.2_{+1.5}$ & $62.2_{+0.7}$ & $67.4_{+0.5}$ & $70.4_{-0.1}$
		\\
		SN-RN18 simple-Ph &
		$59.2_{+3.7}$ & $64.9_{+3.4}$ & $68.3_{+1.4}$ & $71.8_{+1.3}$
		\\
		SN-RN18 simple-TPS &
		$58.2_{+3.1}$ & $63.4_{+1.9}$ & $66.7_{-0.2}$ & $70.1_{-0.4}$
		\\
		SN-RN18 simple-PhTPS &
		$60.8_{+5.3}$ & $64.8_{+3.3}$ & $68.8_{+1.9}$ & $71.1_{+0.6}$
		\\
		\bottomrule
	\end{tabular}
	\label{tab:cs_half_ablation}
\end{table}

\begin{figure}[htb!]
	\centering
	\includegraphics[scale=1]{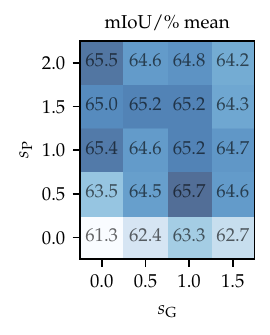}%
	\includegraphics[scale=1]{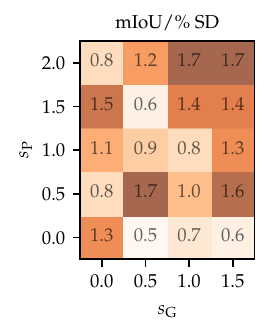}
	\caption{
		Validation of perturbation strength hyper-parameters on Cityscapes val (mIoU/\%). We use 5 different subsets with $1/4$ of the total number of training labels. The hyper-parameters $s_\text{P}$ (photometric) and $s_\text{G}$ (geometric) are defined in the main text. SD denotes the sample standard deviation.}
	\label{fig:cs_half_pert_strength_grid}
\end{figure}

Figure~\ref{fig:cs_half_pert_strength_grid} shows perturbation strength validation using $1/4$ of the labels.
Rows correspond to the factor that multiplies the standard deviation of control point displacements $s_\text{G}$ defined at the end of Section~\ref{subsec:tps}.
Columns correspond to the strength of the photometric perturbation $s_\text{P}$.
The photometric strength $s_\text{P}$ modulates
the random photometric parameters
according to the following expression:
\begin{align}
	(\rvar b, \rvar s, \rvar h, \rvar c) \mapsto (s_\text{P}\cdot\rvar b, \exp(s_\text{P}\cdot \ln(\rvar s)), s_\text{P}\cdot \rvar h, \exp(s_\text{P}\cdot \ln(\rvar c)) \;\text.
\end{align}
We set the probability of choosing a random channel permutation as $\min \{s_\text{P}, 1\}$. Hence, $s_\text{P}=0$ corresponds to the identity function.
Note that the "$1/4$" column in Table~\ref{tab:cs_half_ablation}
uses the same semi-supervised configurations with strengths $s_\text{G}, s_\text{P} \in \cbr{0, 1}$. Also note that the case $(s_\text{G}, s_\text{P}) = (0, 0)$ is slightly different from supervised training in that batch normalization statistics are still updated in the student.
The differences in results are due to variance --
the estimated standard error of the mean of $5$ runs is between $0.35$ and $0.5$.
We can see that the photometric component is more important,
and that a stronger photometric component
can compensate for a weaker geometric component.
Our perturbation strength choice $(s_\text{G}, s_\text{P}) = (1, 1)$
is close to the optimum, which the experiments suggests to be at $(1, 0.5)$.

\begin{figure}[htb!]
	\centering
	\includegraphics[scale=1]{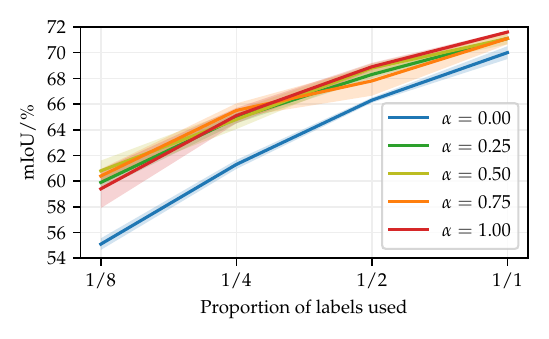}
	\includegraphics[scale=1]{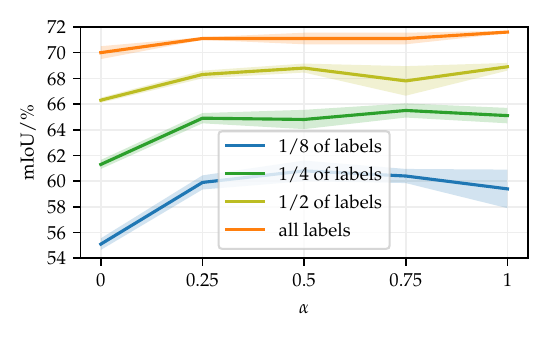}
	\caption{
		Validation of the consistency loss weight $\alpha$ on Cityscapes val (mIoU/\%). We present the same results in two plots with different x-axes: the proportion of labels (left), and the consistency loss weight $\alpha$ (right).}
	\label{fig:cs_half_cons_weight_validation}
\end{figure}

Figure~\ref{fig:cs_half_cons_weight_validation} shows our validation of the consistency loss weight $\alpha$ with SN-RN18 simple-PhTPS. We observe best generalization performance for $\alpha\in\sbr{0.25\bidot 0.75}$. We do not scale the learning rate with $(1+\alpha)^{-1}$ because we use a scale-invariant optimization algorithm.

Appendix~\ref{sec:appendix-additional-experiments}presents experiments that quantify the effect of updating batch normalization statistics when the inputs are perturbed.

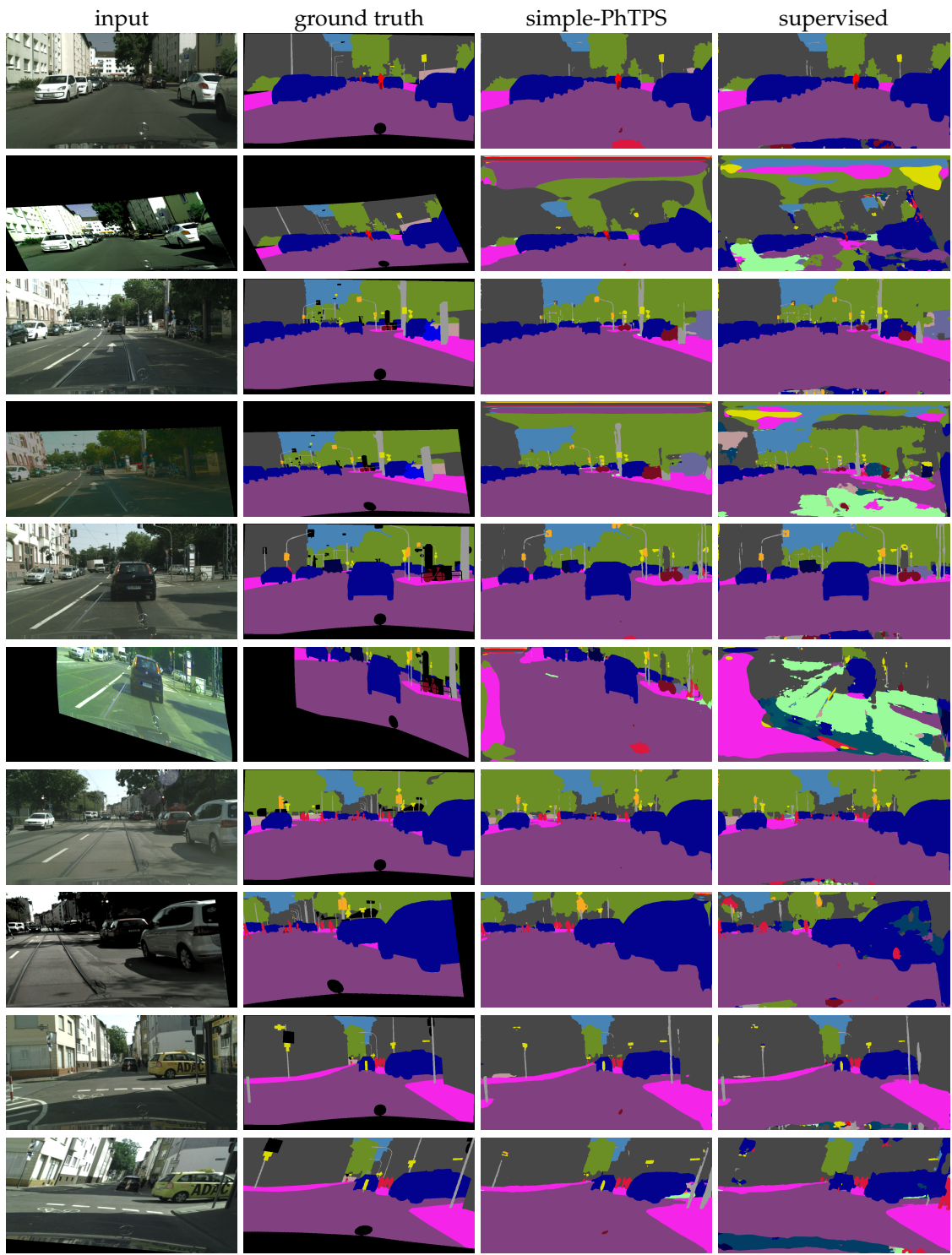
\begin{figure}[htb!]
	\centering
	\input{fig_cs_half_qualitative.tex}
	\caption{
		Qualitative results on the first few validation images with SwiftNet-RN18 trained with $1/4$ of half-resolution Cityscapes labels. Odd rows contain unperturbed inputs, and even rows contain PhTPS perturbed inputs. The columns are (left to right): ground truth segmentations, predictions of simple-PhTPS consistency training, and predictions of supervised training.}
	\label{fig:cs_half_qualitative}
\end{figure}

Figure~\ref{fig:cs_half_qualitative} shows qualitative results
on the first few validation images with SwiftNet-RN18 trained with $1/4$ of labels.
We observe that our method
displays a substantial resilience
to heavy perturbations like those
used during training.

\subsection{Semantic Segmentation on Full-Resolution Cityscapes}

Table~\ref{tab:cs_full} presents our full resolution experiments
in setups like in Table~\ref{tab:cs_half},
and comparison with previous work.
but with full-resolution images and labels.
In comparison with
KE-GAN \cite{qi19cvpr} and
ECS \cite{mendel20eccv},
we underperform with $1/8$ labeled images,
but outperform with $1/2$ labeled images.
Note that KE-GAN \cite{qi19cvpr}
also trains on a large text corpus
(MIT ConceptNet) as well as
that ECS DLv3$^+$-RN50
requires 22\,GiB of GPU memory with batch size 6 \cite{mendel20eccv},
while our SN-RN18 simple-PhTPS
requires less than 8\,GiB of GPU memory
with batch size 8
and can be trained on affordable GPU hardware.
Appendix~\ref{subsec:performance-characteristics} presents
more detailed memory and execution time comparisons with other algorithms.

We note that the concurrent approach DLv3$^+$-RN50 CAC \cite{lai2021cac}
outperforms our method with 1/8 and 1/1 labels.
However, ResNet-18 has significantly
less capacity than ResNet-50.
Therefore, the bottom section
applies our method to the SwiftNet model
with a ResNet-34 backbone,
which still has less capacity than ResNet-50.
The resulting model outperforms DLv3$^+$-RN50 CAC
across most configurations.
This shows that our method
consistently improves
when more capacity is available.

We note that training DLv3$^+$-RN50 CAC
requires three RTX 2080Ti GPUs \cite{lai2021cac},
while our SN-RN34 simple-PhTPS setup
requires less than 9\,GiB of GPU memory
and fits on a single such GPU.
Moreover, SN-RN34 has about 4$\times$ faster inference
than DLv3$^+$-RN50 on RTX 2080Ti.

\begin{table}[htb!]
	\caption{
		Semi-supervised semantic segmentation performance (mIoU/\%)
		on full-resolution Cityscapes val
		with different proportions of labeled data.
		We compare
		simple-PhTPS and MT-PhTPS (ours)
		with supervised training
		and previous work.
		DLv3$^+$-RN50 stands for DeepLab v3$^+$ with ResNet-50,
		and SN for SwiftNet.
		We run experiments
		across 5 different dataset splits
		and report mean mIoUs with standard deviations.
		Best results overall are bold,
		and best results within sections are underlined.
	}
	\centering
	\begin{tabular}{@{\;}l@{\;\;}
			c@{\;\;}c@{\;\;}c@{\;\;}c@{\;}}
		\toprule
		\multirow{2}{*}{\textbf{Method}} & \multicolumn{4}{c}{\textbf{Label proportion}}
		\\
		& $1/8$  &  $1/4$  &  $1/2$  & $1/1$ \\
		\midrule
		KE-GAN \cite{qi19cvpr} &
		$66.9_{\phantom{0.0}}$  & $70.6_{\phantom{0.0}}$ & $72.2_{\phantom{0.0}}$ & $75.3_{\phantom{0.0}}$
		\\
		DLv3$^+$-RN50 supervised \cite{mendel20eccv} &
		$63.2_{\phantom{0.0}}$ & $68.4_{\phantom{0.0}}$ & $72.9_{\phantom{0.0}}$ & $74.8_{\phantom{0.0}}$
		\\
		DLv3$^+$-RN50 ECS \cite{mendel20eccv} &
		$67.4_{\phantom{0.0}}$ & $70.7_{\phantom{0.0}}$ & $72.9_{\phantom{0.0}}$ & $74.8_{\phantom{0.0}}$
		\\
		DLv3$^+$-RN50 supervised \cite{lai2021cac} &
		$63.9_{\phantom{0.0}}$ & $68.3_{\phantom{0.0}}$ & $71.2_{\phantom{0.0}}$ & $76.3_{\phantom{0.0}}$
		\\
		DLv3$^+$-RN50 CAC \cite{lai2021cac} &
		$\mathbf{69.7}_{\phantom{0.0}}$ & $\underline{72.7}_{\phantom{0.0}}$ & ${-}_{\phantom{0.0}}$ & $\underline{77.5}_{\phantom{0.0}}$
		\\
		\midrule
		SN-RN18 supervised &
		$61.1_{0.4}$ & $67.3_{1.1}$ & $71.9_{0.1}$ & $75.4_{0.4}$
		\\
		SN-RN18 simple-PhTPS &
		$66.3_{1.0}$ & $71.0_{0.5}$ & $\underline{74.3}_{0.7}$ & $\underline{75.8}_{0.4}$
		\\
		SN-RN18 MT-PhTPS &
		$\underline{68.6}_{0.6}$ & $\underline{72.0}_{0.3}$ & $73.8_{0.4}$ & $75.0_{0.4}$
		\\
		\midrule
		SN-RN34 supervised &
		$64.9_{0.8}$ & $69.8_{1.0}$ & $73.8_{1.4}$ & $76.1_{0.8}$
		\\
		SN-RN34 simple-PhTPS &
		$69.2_{0.8}$ & $73.1_{0.7}$ & $\mathbf{76.3}_{0.7}$ & $\mathbf{77.9}_{0.2}$
		\\
		SN-RN34 MT-PhTPS &
		$\mathbf{70.8}_{1.5}$ & $\mathbf{74.3}_{0.5}$ & $\mathbf{76.0}_{0.5}$ & $77.2_{0.4}$
		\\
		\bottomrule
	\end{tabular}
	\label{tab:cs_full}
\end{table}

Finally, we present experiments
in the large-data regime
where we place the whole
fine subset into $\set D_\text{l}$.
In some of these experiments we also train
on the large coarsely labeled subset.
We denote extent of supervision
with subscripts "l" (labeled) and "u" (unlabeled). 
Hence, $\mathrm{C_\text{u}}$ in the table
denotes the coarse subset without labels.
Table~\ref{tab:cs-coarse} investigates
the impact of the coarse subset
on SwiftNet performance on
full-resolution Cityscapes val.
We observe that semi-supervised learning
brings considerable improvement
with respect to fully supervised learning
on fine labels only
(columns $\mathrm{F_\text{l}}$ vs
$\mathrm{F_\text{l}} \cup \mathrm{C_\text{u}}$).
It is also interesting to compare
the proposed semi-supervised setup
($\mathrm{F_\text{l}} \cup \mathrm{C_\text{u}}$)
with classic fully supervised
learning on both subsets
($\mathrm{F_\text{l}}, \mathrm{C_\text{l}}$).
We observe that semi-supervised learning
with SwiftNet-RN18
comes close to supervised
learning with coarse labels.
Moreover, semi-supervised learning prevails
when we plug in SwiftNet-RN34.
These experiments suggest that
semi-supervised training represents
an attractive alternative
to coarse labels and large annotation efforts.

\begin{table}[htb!]
	\caption{
		Effects of an additional large dataset on
		supervised and semi-supervised learning
		on full-resolution Cityscapes val (mIoU/\%).
		Tags F and C denote fine and
		coarse subsets, respectively.
		Subset indices denote whether
		we train with labels ($\textrm{l}$)
		or one-way consistency ($\textrm{u}$).
	}
	\centering
	\begin{tabular}{@{\;}l@{\;\;}c@{\;\;}c@{\;\;}c@{\;\;}c@{\;\;}c@{\;}}
		\toprule
		\textbf{Method} &
		\multicolumn{1}{c}{$\mathrm{F_\text{l}}$} &
		\multicolumn{1}{c}{$(\mathrm{F_\text{l}},\mathrm{F_\text{u}})$} &
		\multicolumn{1}{c}{$(\mathrm{F_\text{l}},
			\mathrm{F_\text{u}}\cup\mathrm{C_\text{u}})$} &
		\multicolumn{1}{c}{$(\mathrm{F_\text{l}},
			\mathrm{C_\text{l}})$} \\
		\midrule
		SN-RN18 simple-PhTPS &
		\multirow{2}{*}{$75.4_{0.4}$} & $75.8_{0.4}$ & $76.5_{0.3}$ & \multirow{2}{*}{$\mathbf{76.9}_{0.3}$} \\
		SN-RN18 MT-PhTPS &
		& $75.0_{0.4}$ & $75.5_{0.3}$ & \\
		\midrule
		SN-RN34 simple-PhTPS &
		$76.1_{0.8}$ & $77.9_{0.2}$ & $\mathbf{78.5}_{0.4}$ & $77.7_{0.4}$ \\  
		\bottomrule
	\end{tabular}
	\label{tab:cs-coarse}
\end{table}

\subsection{Validation of Consistency Variants}

Table~\ref{tab:cons-variants} presents experiments with
supervised baselines and four variants of semi-supervised consistency training.
All semi-supervised experiments use the same PhTPS perturbations
on CIFAR-10 (4000 labels and 50000 images)
and half-resolution Cityscapes (the SwiftNet-RN18 setups with $1/4$ labels from Table~\ref{tab:cs_half}).
We investigate the following kinds of consistency:
one-way with clean teacher (1w-ct, cf. Figure~\ref{fig:moons-1w-ct}),
one-way with clean student (1w-cs, cf. Figure~\ref{fig:moons-1w-cs}),
two-way with one clean input (2w-c1, cf. Figure~\ref{fig:moons-2w-c1}), and
one-way with both inputs perturbed (1w-p2).
Note that two-way consistency is not
possible with Mean Teacher.
Also, when both inputs are perturbed (1w-p2),
we have to use the inverse geometric transformation
on dense predictions \cite{bortsova19miccai}.
We achieve that by forward warping \cite{niklaus2020softsplat}
with the same displacement field.
Two-way consistency with both inputs perturbed (2w-p2)
is possible as well.
We expect it to behave similarly to 1w-2p
because it could be seen as a superposition
of two opposite one-way consistencies,
and our preliminary experiments suggest so.

We observe that 1w-ct outperforms all other variants,
while 2w-c1 performs in-between 1w-ct and 1w-cs.
This confirms our hypothesis that predictions in clean inputs
make better consistency targets.
We note that 1w-p2 often outperforms 1w-cs,
while always underperforming with respect to 1w-ct.
A closer inspection suggests
that 1w-p2 sometimes learns
to cheat the consistency loss
by outputting similar predictions
for all perturbed images.
This occurs more often when
batch normalization uses the batch statistics
estimated during training.
A closer inspection of 1w-cs experiments on Cityscapes
indicates consistency cheating combined with
severe overfitting to the training dataset.

\begin{table}[htb!]
	\caption{
		Comparison of 4 consistency variants under
		PhTPS perturbations:
		one-way with clean teacher (1w-ct),
		one-way with clean student (1w-cs),
		two-way with one input clean (2w-c1), and
		one-way with both inputs perturbed (1w-p2).
		Algorithms are evaluated on
		CIFAR-10 test (accuracy/\%)
		while training on 4000 out of 50000 labels (CIFAR-10, $2/25$)
		and half-resolution Cityscapes val (mIoU/\%)
		while training on $1/4$ of labels
		from Cityscapes train
		with SwiftNet-RN18 (CS-half, $1/4$).
	}
	\centering
	\begin{tabular}{@{}l@{\;\;}l@{\;\;}|c@{\;\;}
			c@{\;\;}c@{\;\;}c@{\;\;}c@{}}
		\toprule
		\textbf{Dataset} & \textbf{Method} & {sup.} &  {1w-ct} &
		{1w-cs} & {2w-c1} & {1w-p2} \\
		\midrule
		CIFAR-10, 4k & WRN-28-2 simple-PhTPS & $80.8_{0.4}$ & $\mathbf{90.8}_{0.3}$ & $69.3_{4.2}$ & $72.9_{2.6}$ & $73.3_{7.0}$ \\
		CIFAR-10, 4k & WRN-28-2 MT-PhTPS & $80.8_{0.4}$ & $\mathbf{90.8}_{0.4}$ & $80.5_{0.5}$ & - & $73.4_{1.4}$ \\
		CS-half, 1/4 & SN-RN18 simple-PhTPS & $61.5_{0.5}$ & $\mathbf{65.3}_{1.9}$ & $\phantom{0}1.6_{1.0}$ & $16.7_{3.0}$ & $61.6_{0.5}$ \\
		CS-half, 1/4 & SN-RN18 MT-PhTPS & $61.5_{0.5}$ & $\mathbf{66.0}_{1.0}$ & $61.5_{1.4}$ & - & $62.0_{1.1}$ \\
		\bottomrule
	\end{tabular}
	\label{tab:cons-variants}
\end{table}

%

\subsection{Image Classification on CIFAR-10}

Table~\ref{tab:cifar10} evaluates image classification performance
of two supervised baselines
and 4 semi-supervised algorithms on CIFAR-10.
The first supervised baseline uses only labeled data
with standard data augmentation.
The second baseline
additionally uses our perturbations
for data augmentation.
The third algorithm is VAT
with entropy minimization \cite{miyato19pami}.
The simple-PhTPS approach outperforms
supervised approaches and VAT.
Again, two-way consistency
results in the worst generalization performance.
Perturbing the teacher input
results in accuracy below 17\%
for 4000 or less labeled examples,
and is not displayed.
Note that somewhat better performance can be achieved
by complementing consistency with other techniques
that are either unsuitable for dense prediction or out of scope of this paper \cite{cubuk20nips,xie20nips,berthelot19nips}.

\begin{table}[htb!]
	\caption{
		Classification accuracy [\%]
		on CIFAR-10 test with WRN-28-2.
		We compare two supervised approaches (top),
		VAT with entropy minimization \cite{miyato19pami}
		(middle),
		and two-way and one-way consistency
		with our perturbations
		(bottom three rows).
		We report means and standard deviations of $5$ runs.
		The label "supervised PhTPS-aug" denotes
		supervised training where
		half of each mini-batch
		is perturbed with PhTPS.
	}
	\centering
	\begin{tabular}
		{@{\,}l@{\;\;}c@{\;\;}c@{\;\;}c@{\;\;}c@{}}
		\toprule
		\multirow{2}{*}{\textbf{Method}} & \multicolumn{4}{@{}c@{}}
		{\textbf{Number of labeled examples}}
		\\
		&  {250}  &  {1000}  &  {4000}  & {50000}
		\\
		\midrule
		supervised
		& $31.8_{0.6}$ & $59.3_{1.4}$
		& $81.0_{0.2}$ & $94.7$ 
		\\
		supervised PhTPS-aug
		& $48.7_{0.9}$ & $67.2_{0.5}$
		& $81.7_{0.2}$ & $95.0$ 
		\\
		\midrule
		VAT + entropy minimization
		& $41.0_{2.5}$ & $73.2_{1.5}$
		& $84.2_{0.4}$ & $90.5_{0.2}$
		\\
		1w-cs simple-PhTPS
		& $27.7_{5.9}$ & $51.7_{3.5}$
		& $69.3_{4.2}$ & $91.6_{1.5}$
		\\
		2w-c1 simple-PhTPS
		& $30.3_{1.8}$ & $54.8_{1.5}$
		& $72.9_{2.6}$ & $95.9_{0.2}$
		\\
		1w-ct simple-PhTPS
		& $\mathbf{68.8}_{5.4}$ & $\mathbf{84.2}_{0.4}$
		& $\mathbf{90.6}_{0.4}$ & $\mathbf{96.2}_{0.2}$
		\\
		\bottomrule
	\end{tabular}
	\label{tab:cifar10}
\end{table}

\section{Discussion} \label{sec:conclusion}

We have presented the first comprehensive study
of one-way consistency for semi-supervised dense prediction,
and proposed a novel perturbation model
which leads to competitive generalization performance on Cityscapes.
Our study clearly shows that
one-way consistency with clean teacher
outperforms other forms of consistency
(e.g.~clean student or two-way)
both in terms of generalization performance
and training footprint.
We explain this by observing
that predictions in perturbed images
tend to be less reliable targets.

The proposed perturbation model is a composition
of a photometric transformation and a geometric warp.
These two kinds of perturbations
have to be treated differently since we desire
invariance to the former
and equivariance to the latter.
Our perturbation model outperforms CutMix
both in standard experiments with DeepLabv2-RN101
and in combination with
recent efficient models (SwiftNet-RN18 and SwiftNet-RN34).

We consider two teacher formulations.
In the simple formulation the teacher
is a frozen copy of the student.
In the Mean Teacher formulation,
the teacher is a moving average of student parameters.
Mean Teacher outperforms simple consistency
in low data regimes (half resolution, few labels).
However, experiments with more data
suggest that the simple one-way formulation
scales significantly better.

To the best of our knowledge,
this is the first account of
semi-supervised semantic segmentation
with efficient models.
This combination is essential
for many practical real-time applications
where there is a lack of large datasets
with suitable pixel-level groundtruth.
Many of our experiments are based on SwiftNet-RN18,
which behaves similarly to DeepLabv2-RN101,
while offering
about 9$\times$ faster inference on half-resolution images,
and about 15$\times$ faster inference on full-resolution images
on RTX 2080Ti.
Experiments on Cityscapes coarse
reveal that semi-supervised learning
with one-way consistency
can come close and exceed
full supervision with coarse annotations.
Simplicity, competitive performance and speed of training
make this approach a very attractive baseline
for evaluating future semi-supervised approaches
in the dense prediction context.





\vspace{6pt}

\bibliographystyle{plain}
{\small
	\bibliography{bibliography}
}

\appendix

\section[\appendixname~\thesection]{Additional Algorithm Comparisons}\label{sec:appendix-comparisons}

\subsection[\appendixname~\thesubsection]{Hyper-Parameters} \label{subsec:hyper-parameter-overview}

{Tables~\ref{tab:hyper-parameters} and \ref{tab:optimizer-hyper-parameters} review hyper-parameters of consistency-based semi-supervised algorithms from Tables~\ref{tab:cs_half} and ~\ref{tab:cs_full}.}

\newgeometry{margin=2cm}
\startlandscape
\thispagestyle{empty}

\begin{table}[!htb]
	\caption{Overview of hyper-parameters of consistency-based algorithms for semi-supervised semantic segmentation. We denote the setup that we use in our experiments including supervised training with ``our''. We use the following symbols: $\vec y$ is the teacher prediction, $\tilde{\vec y}$ is the student prediction, $\mathrm{hard}$ is a function that maps a vector representing a distribution to the closest one-hot vector ($\mathrm{hard}_\ind{c}=\enbbracket{c = \argmax_k \vec y_\ind{k}}$), $e$ is the proportion of completed epochs, $\alpha$ is the consistency loss weight, and $\eta$ is the base learning rate. The if-clause in the ``Consistency loss'' column represents confidence thresholding, i.e., determines whether the pixel is included in loss computation.}
	\centering
	\begin{tabular}{llrcrrrrcrc}
		\toprule
		\textbf{Model} & \textbf{Method}                                             & \textbf{Crop Size} & \textbf{Jitter. Scale}         & \textbf{Iterations} & \textbf{Epochs} & $B_l$ & $B_u$ & \textbf{Consistency Loss}                                          & $\alpha$         & \textbf{Learning Rate}      \\
		\midrule
		DLv2  & MT-CutMix \cite{french20bmvc}                      & 321       & $\sbr{0.5\bidot 1.5}$ & 40,000      & 135 * & 10    & 10    & $\enVert{\mathrm{hard}(\vec y)-\tilde{\vec y}}_2^2$ if $\max_c \vec y_\ind{c}>0.97$ & 0.5                      & $\eta(1-e)^{0.9}$  \\
		DLv2  & MT-CutMix\textsubscript{$\sim$\cite{french20bmvc}} & 321       & $\sbr{0.5\bidot 1.5}$ & 37,100      & 100    & 4     & 4     & $\enVert{\mathrm{hard}(\vec y)-\tilde{\vec y}}_2^2$ if $\max_c \vec y_\ind{c}>0.97$
		& 1                        & $\eta(1-e)^{0.9}$  \\
		DLv2  & ours                                                & 448       & $\sbr{0.5\bidot 2}$   & 74,200      & 200    & 4     & 4     & $\mathit{D}(\vec y, \tilde{\vec y})$                      & 0.5                      & $\eta(1-e)^{0.9}$  \\
		SN    & ours                                                & 448       & $\sbr{0.5\bidot 2}$   & 74,200      & 200    & 8     & 8     & $\mathit{D}(\vec y, \tilde{\vec y})$                      & 0.5                      & $\eta\cos(e\pi/2)$ \\
		SN    & ours                                                & 768       & $\sbr{0.5\bidot 2}$   & 92,750      & 250    & 8     & 8     & $\mathit{D}(\vec y, \tilde{\vec y})$                      & 0.5                      & $\eta\cos(e\pi/2)$ \\
		DLv3$^+$ & CAC \cite{lai2021cac}                              & 720       & $\sbr{0.5\bidot 1.5}$ & 92,560      & 249 *    & 8     & 8     & see \cite{lai2021cac}                                                  & $\enbbracket{e>5/80}\cdot0.1$ & $\eta(1-e)^{0.9}$  \\
		DLv3$^+$ & ours                                                & 720       & $\sbr{0.5\bidot 1.5}$ & 92,560      & 249 *    & 8     & 8     & $\mathit{D}(\vec y, \tilde{\vec y})$                      & 0.5                      & $\eta(1-e)^{0.9}$ \\
		\bottomrule
	\end{tabular}
		\\\noindent{\footnotesize{$^*$ Some authors \cite{french20bmvc, lai2021cac} use the word ``epoch'' to refer to a fixed number of iterations --- 1000 and 1157 iterations, respectively.}}
	\label{tab:hyper-parameters}
\end{table}

\begin{table}[!htb]
	\caption{Optimizer hyper-parameter configurations
			for Cityscapes semantic segmentation experiments, represented in a PyTorch-like style.}
	\centering
	\begin{tabular}{llll}
			\toprule
			\multirow{2}{*}{\textbf{Model}} & \multirow{2}{*}{\textbf{Method}} & \multicolumn{2}{c}{\textbf{Optimizer Configuration}} \\
			& & \textbf{Main} & \textbf{Backbone (Difference)} \\
			\midrule
			DLv2 & MT-CutMix \cite{french20bmvc}
			& \texttt{SGD, lr=$3\cdot 10^{-5}$, momentum=$0.9$, weight\_decay=$5\cdot 10^{-4}$}
			& \texttt{} \\
			DLv2 & MT-CutMix\textsubscript{$\sim$\cite{french20bmvc}}
			& \texttt{SGD, lr=$3\cdot 10^{-5}$, momentum=$0.9$, weight\_decay=$5\cdot 10^{-4}$}
			& \texttt{} \\
			DLv2 & ours
			& \texttt{Adam, betas=$(0.9, 0.99)$, lr=$4\cdot 10^{-4}$, weight\_decay=$1\cdot 10^{-4}$}
			& \texttt{lr=$1\cdot 10^{-4}$, weight\_decay=$2.5\cdot 10^{-5}$}           \\
			SN & ours
			& \texttt{Adam, betas=$(0.9, 0.99)$, lr=$4\cdot 10^{-4}$, weight\_decay=$2.5\cdot 10^{-5}$}
			& \texttt{lr=$1\cdot 10^{-4}$, weight\_decay=$6.25\cdot 10^{-6}$} \\
			DLv3$^+$ & CAC \cite{lai2021cac}
			& \texttt{SGD, lr=$1\cdot 10^{-1}$, momentum=$0.9$}
			& \texttt{lr=$1\cdot 10^{-2}$, weight\_decay=$1\cdot 10^{-4}$} \\
			DLv3$^+$ & ours
			& \texttt{SGD, lr=$1\cdot 10^{-1}$, momentum=$0.9$}
			& \texttt{lr=$1\cdot 10^{-2}$,  weight\_decay=$1\cdot 10^{-4}$} \\
			\bottomrule
	\end{tabular}
	\label{tab:optimizer-hyper-parameters}
\end{table}
\finishlandscape
\restoregeometry

\subsection[\appendixname~\thesubsection]{Time and Memory Performance Characteristics} \label{subsec:performance-characteristics}

Table~\ref{tab:method-speeds} shows memory requirements and training times of methods from Tables~\ref{tab:cs_half} and \ref{tab:cs_full}. The times include data loading and processing, and do not include evaluation on the validation set. The memory measurements are based on the \texttt{max\_memory\_allocated} \footnote{Note that some overhead that is required by PyTorch is not included in this measurement. See PyTorch memory management documentation for more information: \url{https://pytorch.org/docs/master/notes/cuda.html}.} and \texttt{reset\_peak\_memory\_stats} procedures from the \texttt{torch.cuda} package. Some algorithms did not fit into the memory of GTX 2080Ti.
The memory allocations are higher than in Figure~\ref{fig:cs_memory_alloc} because the supervised prediction, perturbed inputs, and perturbed outputs are unnecessarily kept in memory.

For DeepLabv3$^+$-RN50, we use the number of iterations, batch size, and crop size from \cite{lai2021cac}. Note, however, that the method from \cite{lai2021cac} has the memory requirements of two-way consistency because of per-pixel directionality.

\begin{table}[!htb]
\caption{Half resolution Cityscapes (top section) and Cityscapes (bottom section) maximum memory allocation and training time on two GPUs.}
\centering
\begin{tabular}{llrrrr|rrr}
	\toprule
    \multirow{2}{*}{\textbf{Model}} & \multirow{2}{*}{\textbf{Method}} & \multirow{2}{*}{\shortstack[r]{\textbf{Crop}\\\textbf{Size}}} & \multirow{2}{*}{\textbf{Iterations}} & \multirow{2}{*}{$B_l$} & \multirow{2}{*}{$B_u$} & \multirow{2}{*}{\shortstack[r]{\textbf{Memory}\\\textbf{$ / \mathrm{MiB}$}}} & \multicolumn{2}{c}{\textbf{Duration$ / \mathrm{min}$}} \\
	& & & & & & &  \textbf{A4500}     & \textbf{2080Ti} \\
	\midrule
	DLv2-RN101 & MT-CutMix \cite{french20bmvc} & 321 & 40000 & 10 & 10 & 16289 & 1067 & -- \\
	& MT-CutMix\textsubscript{$\sim$\cite{french20bmvc}} & 321 & 37100 & 4 & 4 & 7037 & 794 & 1314 \\
	DLv2-RN101 & supervised & 448 & 74300 & 4 & -- & 6611 & 338 & 602 \\
	& MT-PhTPS & 448 & 74300 & 4 & 4 & 7021 & 816 & 1397 \\
	SN-RN18 & supervised & 448 & 74200 & 4 & -- & 1646 & 119 & 161 \\
	& simple-PhTPS & 448 & 74200 & 8 & 8 & 2398 & 228 & 279 \\
	& MT-PhTPS & 448 & 74200 & 8 & 8 & 2456 & 234 & 297 \\
	\midrule
	SN-RN18 & supervised & 768 & 92750 & 8 & -- & 4444 & 321 & 432 \\
	& simple-PhTPS & 768 & 92750 & 8 & 8 & 6683 & 732 & 963 \\
	& MT-PhTPS & 768 & 92750 & 8 & 8 & 6727 & 768 & 965 \\
	SN-RN34 & supervised & 768 & 92750 & 8 & -- & 5500 & 422 & 570 \\
	& simple-PhTPS & 768 & 92750 & 8 & 8 & 7737 & 994 & 1268 \\
	& MT-PhTPS & 768 & 92750 & 8 & 8 & 7818 & 1013 & 1276 \\
	DLv3+-RN50 & supervised & 720 & 92560 & 8 & -- & 11645 & 1229 & -- \\
	& simple-PhTPS & 720 & 92560 & 8 & 8 & 13384 & 1884 & -- \\
	& CAC \cite{lai2021cac} & 720 & 92560 & 8 & 8 & 25165$^\dagger$ & $>$3000$^*$ & --  \\
	\bottomrule
\end{tabular}
	\\\noindent{\footnotesize{
	$^\dagger$ The original implementation requires 36005\,MiB. Approximately 10.6\,GiB can be saved by accumulating gradients as in Algorithm~\ref{alg:training}.
	\\
	$^*$ Estimated by running on NVidia A100.
	\\}}
\label{tab:method-speeds}
\end{table}

Table~\ref{tab:model-parameters} shows the numbers of model parameters, and Table~\ref{tab:model-speeds} shows inference speeds of models from Tables~\ref{tab:cs_half} and \ref{tab:cs_full}.

\begin{table}[!htb]
\caption{Number of model parameters.}
\centering
\begin{tabular}{lr}
\toprule
\textbf{Model}         & \textbf{Number of parameters} \\
\midrule
DeepLabv2-RN101    & $43.80\cdot 10^6$                               \\
DeepLabv3$^+$-RN50   & $40.35\cdot 10^6$                               \\
SwiftNet-RN34 & $21.91\cdot 10^6$                               \\
SwiftNet-RN18 & $11.80\cdot 10^6$                               \\
\bottomrule
\end{tabular}
\label{tab:model-parameters}
\end{table}

\begin{table}[!htb]
\caption{Model inference speed (number of iterations per second) on three different GPUs and two input resolutions. Inputs are processed one by one, without overlap in computation. The measurements include the computation of the cross-entropy loss, but do not include data loading and preparation.}
\centering
\begin{tabular}{lrrrrrr}
	\toprule
	& \multicolumn{3}{c}{\boldmath{$1024\times2048$}} & \multicolumn{3}{c}{\boldmath{$512\times 1024$}} \\
	& \textbf{A4500}      & \textbf{2080Ti }    & \textbf{1080Ti}     & \textbf{A4500}      & \textbf{2080Ti}     & \textbf{1080Ti}     \\
	\midrule
	DeepLabv2-RN101    & 5.1        & 3.0        & 1.5        & 19.6       & 12.2       & 6.3        \\
	DeepLabv3+-RN50   & 16.1       & 9.6        & 5.2        & 54.2       & 30.7       & 23.5       \\
	SwiftNet-RN34 & 39.2       & 30.5       & 23.6       & 93.4       & 86.1       & 73.5       \\
	SwiftNet-RN18 & 56.5       & 45.3       & 34.6       & 139.5      & 115.8      & 98.4       \\
	\bottomrule
\end{tabular}
\label{tab:model-speeds}
\end{table}

\section[\appendixname~\thesection]{Effect of Updating Batch Normalization Statistics in the Perturbed Student}\label{sec:appendix-additional-experiments}

Batch normalization layers estimate
population statistics during training
as exponential moving averages.
Training on perturbed images
can adversely affect suitability
of these estimates.
Consequently, we explore the design choice
to update the statistics only on clean inputs
(when the supervised loss is computed),
instead of both on clean and on perturbed inputs.

\begin{figure}[htb!]
\centering
\begin{subfigure}[t]{1\linewidth}
\centering
\includegraphics[scale=1]{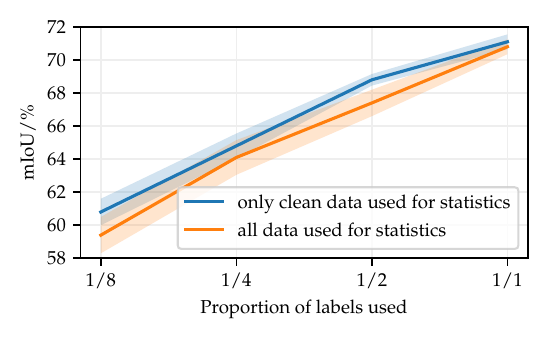}
\caption{Half-resolution Cityscapes val.}
\label{fig:cs_half_ablation_bn_stat_upd}
\end{subfigure}
\\\bigskip
\begin{subfigure}[t]{1\linewidth}
\centering
\includegraphics[scale=1]{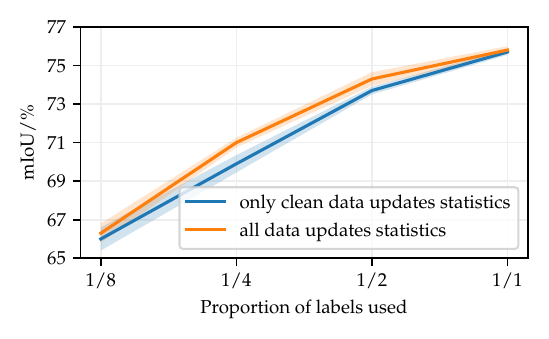}
\caption{Cityscapes val.}
\label{fig:cs_ablation_bn_stat_upd}
\end{subfigure}
\\\bigskip
\begin{subfigure}[t]{1\linewidth}
\centering
\includegraphics[scale=1]{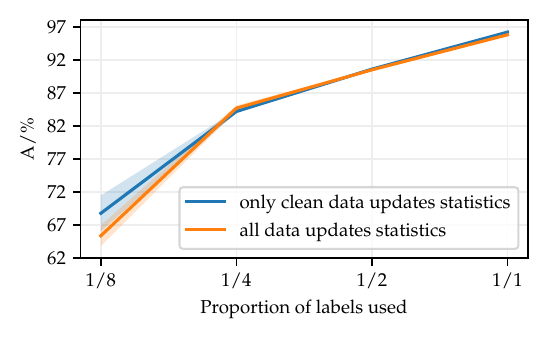}
\caption{CIFAR-10 validation set.}
\label{fig:cifar10_half_ablation_bn_stat_upd}
\end{subfigure}
\\\bigskip
\caption{
Effect of updating batch normalization statistics in the perturbed student. }
\label{fig:ablation_bn_stat_upd}
\end{figure}

Note that this configuration difference does not affect parameter optimization because the training always relies on mini-batch statistics..
Figure~\ref{fig:ablation_bn_stat_upd} shows the effect of disabling the updates of batch normalization statistics when the model instance (student) receives perturbed inputs in our semi-supervised training (one way consistency with clean teacher). The experiments are conducted according to the corresponding descriptions in section~\ref{sec:results}.
In case of half-resolution Cityscapes, disabling updates in the perturbed student (blue) increased the validation mIoU by between $0.3$ and $1.4$\,pp, depending on the proportion of labels. However, in case of full-resolution Cityscapes, an opposite effect occured -- mIoU decreased by between $0.1$ and $1.1$\,pp.
In CIFAR-10 experiments, the effect is mostly neutral.

\end{document}

%% file: math.tex
\usepackage{amsmath}
\usepackage{amssymb}  

\usepackage{mathtools}  
\usepackage{bm}

\DeclareMathOperator{\dif}{d\!}

\makeatletter
\newcommand{\spx}[1]{%
	\if\relax\detokenize{#1}\relax
	\expandafter\@gobble
	\else
	\expandafter\@firstofone
	\fi
	{^{#1}}%
}
\makeatother

\newcommand\pd[3][]{\frac{\partial\spx{#1}#2}{\partial#3\spx{#1}}}

\newcommand{\genericdel}[4]{%
	\ifcase#3\relax
	\ifx#1.\else#1\fi#4\ifx#2.\else#2\fi\or
	\bigl#1#4\bigr#2\or
	\Bigl#1#4\Bigr#2\or
	\biggl#1#4\biggr#2\or
	\Biggl#1#4\Biggr#2\else
	\left#1#4\right#2\fi
}
\newcommand{\del}[2][-1]{\genericdel(){#1}{#2}}
\newcommand{\cbr}[2][-1]{\genericdel\{\}{#1}{#2}}
\let\set\cbr

\newcommand{\sbr}[2][-1]{\genericdel[]{#1}{#2}}

\newcommand{\intcc}[2][-1]{\mathinner{\genericdel[]{#1}{#2}}}

\newcommand{\envert}[2][-1]{\genericdel\vert\vert{#1}{#2}}

\newcommand{\enVert}[2][-1]{\genericdel\|\|{#1}{#2}}
\let\norm\enVert

\newcommand{\ind}[1]{{\sbr{#1}}}

\usepackage{stmaryrd}  

\usepackage[OMLmathsfit]{isomath}  
\usepackage{upgreek}

\DeclareMathAlphabet{\mathbbmsl}{U}{bbm}{m}{sl}
\DeclareMathAlphabet{\mathbbmb}{U}{bbm}{b}{it}
\DeclareMathAlphabet{\mathbbmssit}{U}{bbmss}{m}{it}

\let\vec\relax
\let\set\relax
\newcommand{\vec}[1]{\bm{#1}}
\newcommand{\set}[1]{\mathbbmsl{#1}}

\newcommand{\cvec}[1]{\mathbf{#1}}
\newcommand{\cset}[1]{\mathbb{#1}}

\newcommand{\nunder}[2][5]{\mathrlap{\mkern\the\numexpr#1/2mu\relax\underline{\phantom{\mathrm{#2}\mkern-#1mu}}}\mathrm{#2}}
\newcommand{\subline}[2][n]{%
    \makebox[\widthof{$#2$}/2]{}\mathclap{#2}%
    \text{\smash{\raisebox{0.04em}{
        $\mathclap{\underline{\hphantom{#1}\vphantom{#2}}}$%
    }}}%
    \makebox[\widthof{$#2$}/2]{}%
}

\newcommand{\rvarstyle}[1]{{{\subline[i]{#1}}}}
\newcommand{\rvar}[1]{\rvarstyle{#1}}
\newcommand{\rvec}[1]{\rvarstyle{\vec{#1}}}

\newcommand{\transpose}{\mathsf T}
\newcommand{\tp}{\transpose}

\DeclareMathOperator{\dom}{dom}

\DeclareMathOperator*{\argmax}{arg\,max}
\let\P\relax
\DeclareMathOperator{\P}{P}
\let\p\relax
\DeclareMathOperator{\p}{\mathrm{p}}
\DeclareMathOperator*{\E}{\mathrm{I\kern-.282em E}}
\DeclareMathOperator*{\D}{\mathrm{I\kern-.282em D}}

\newcommand{\enbbracket}[1]{{\mathinner{\left\llbracket{#1}\right\rrbracket}}}

\let\oldhat\hat
\renewcommand{\hat}[1]{\vphantom{#1}\smash[t]{\oldhat{#1}}}
\let\oldtilde\tilde
\renewcommand{\tilde}[1]{\vphantom{#1}\smash[t]{\oldtilde{#1}}}
\let\oldwidetilde\widetilde
\renewcommand{\widetilde}[1]{\vphantom{#1}\smash[t]{\oldwidetilde{#1}}}

\newcommand{\bidot}{\mkern1.5mu{..}\mkern1.5mu}

\newenvironment{talign*}
{\csname align*\endcsname}
{\endalign}

\usepackage{dashbox}%

%% file: fig_cs_half_qualitative.tex
\newcommand{\imwidth}{0.24\linewidth}
\begin{tabular}{@{}c@{\;}c@{\;}c@{\;}c@{}}
	input & ground truth & simple-PhTPS & supervised \\
	\includegraphics[width=\imwidth]{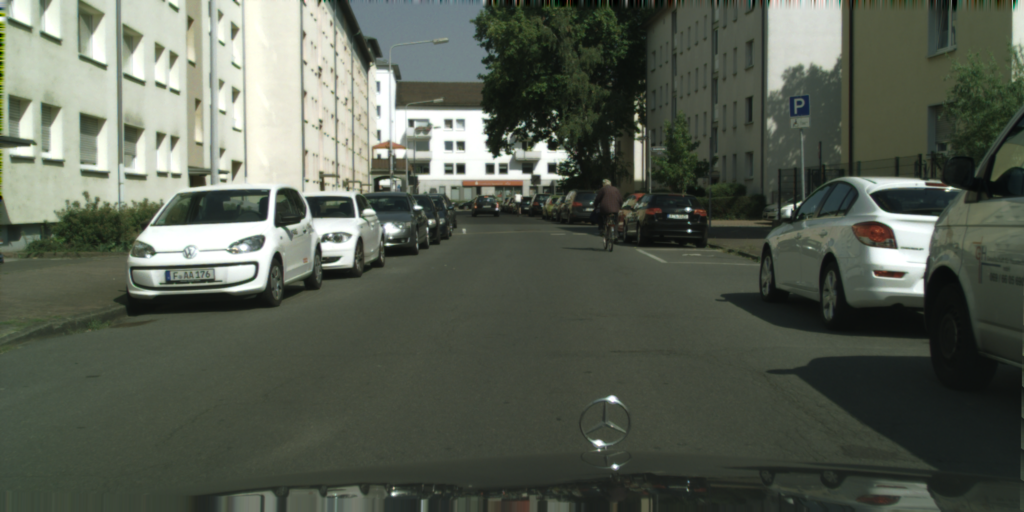}&
	\includegraphics[width=\imwidth]{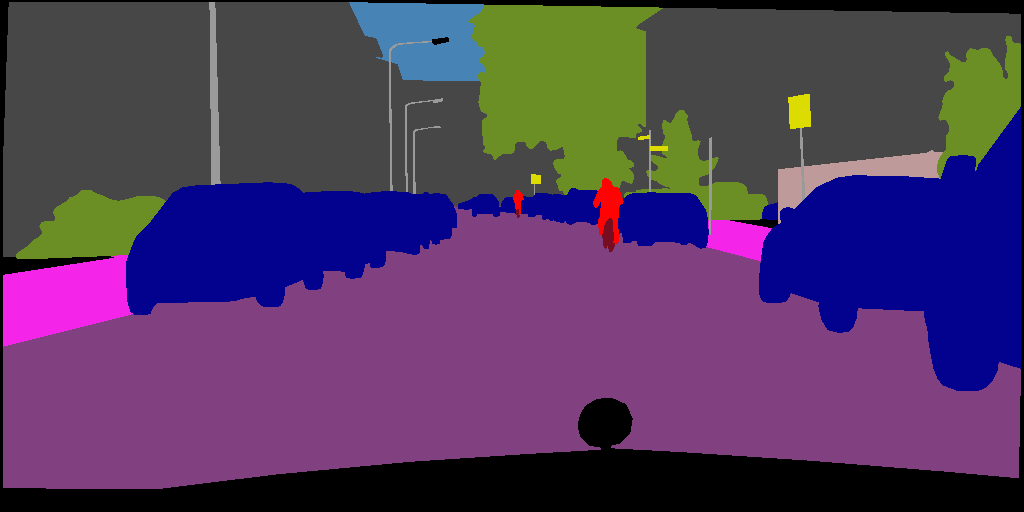}&
	\includegraphics[width=\imwidth]{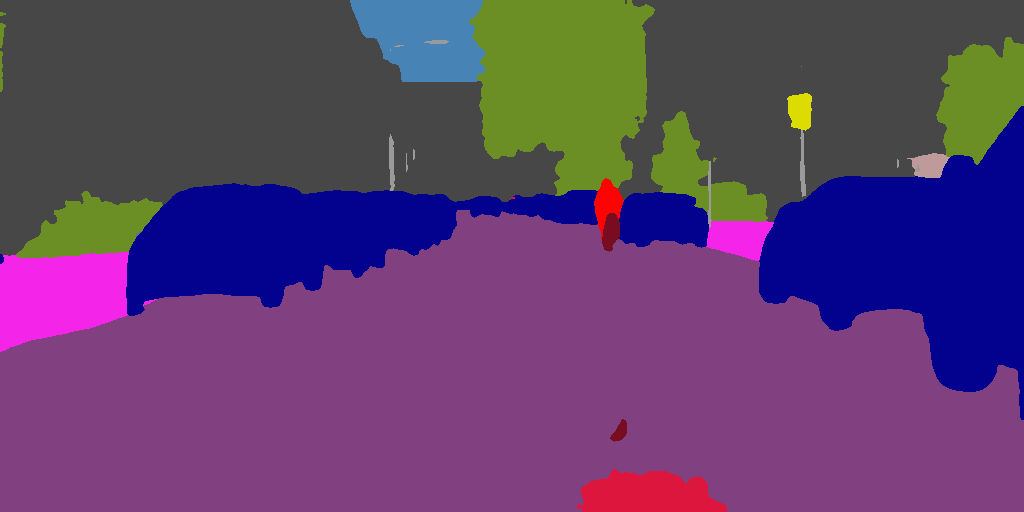}&
	\includegraphics[width=\imwidth]{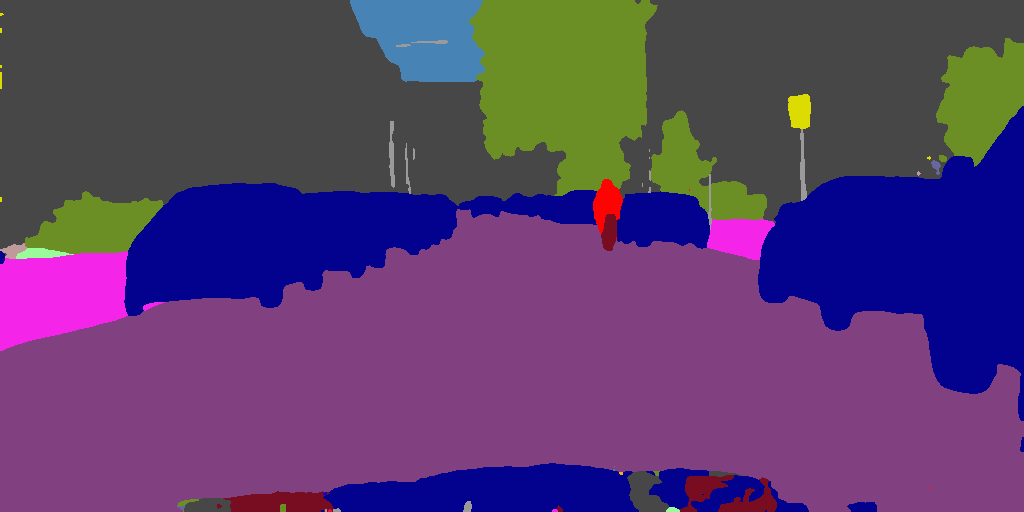}\\
	\includegraphics[width=\imwidth]{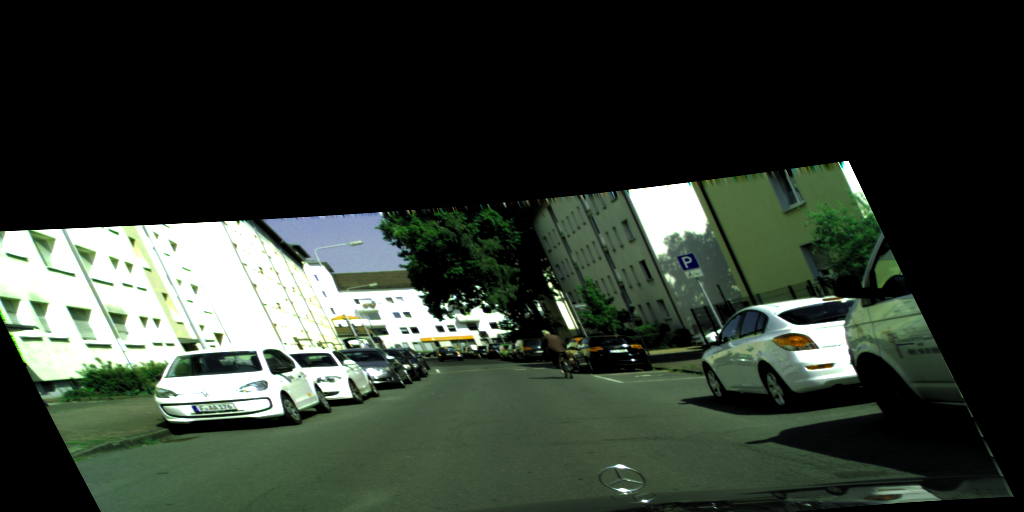}&
	\includegraphics[width=\imwidth]{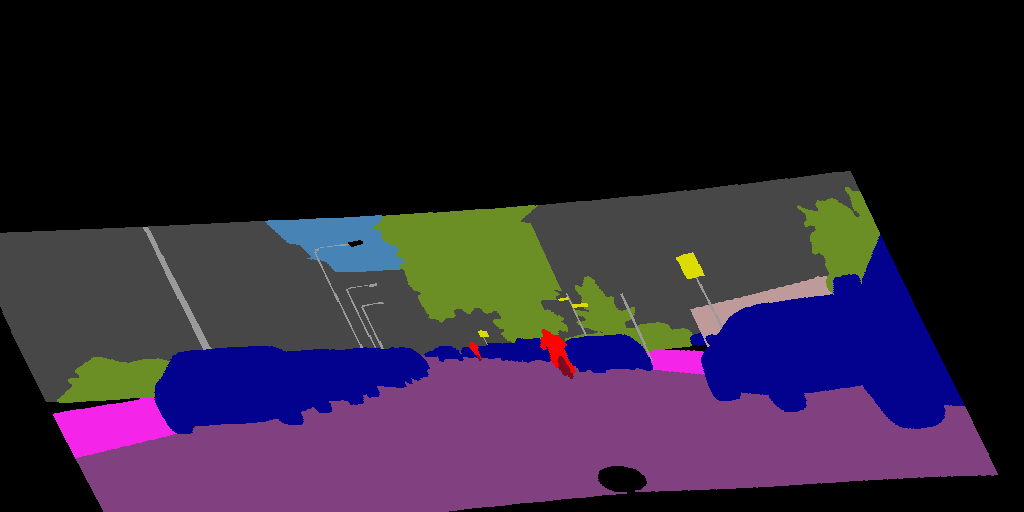}&
	\includegraphics[width=\imwidth]{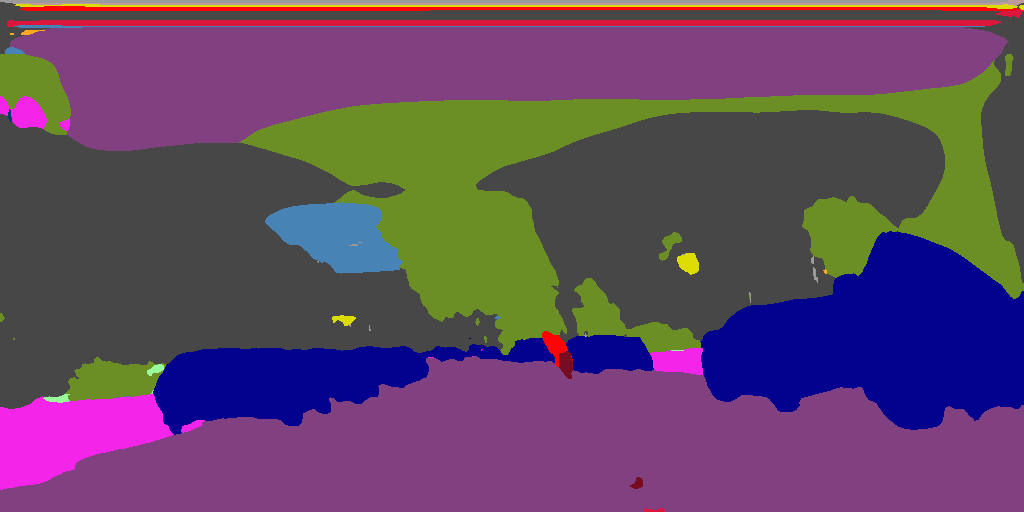}&
	\includegraphics[width=\imwidth]{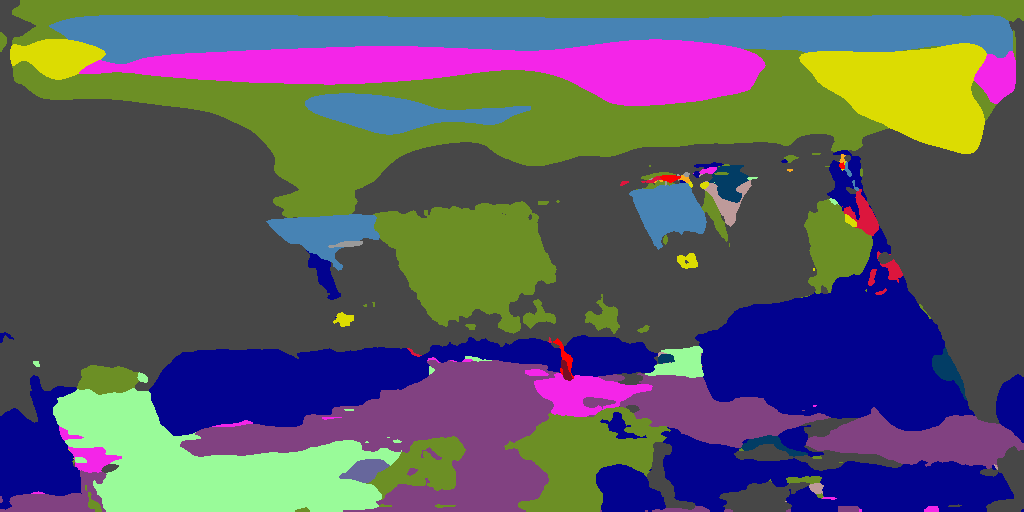}\\
	\includegraphics[width=\imwidth]{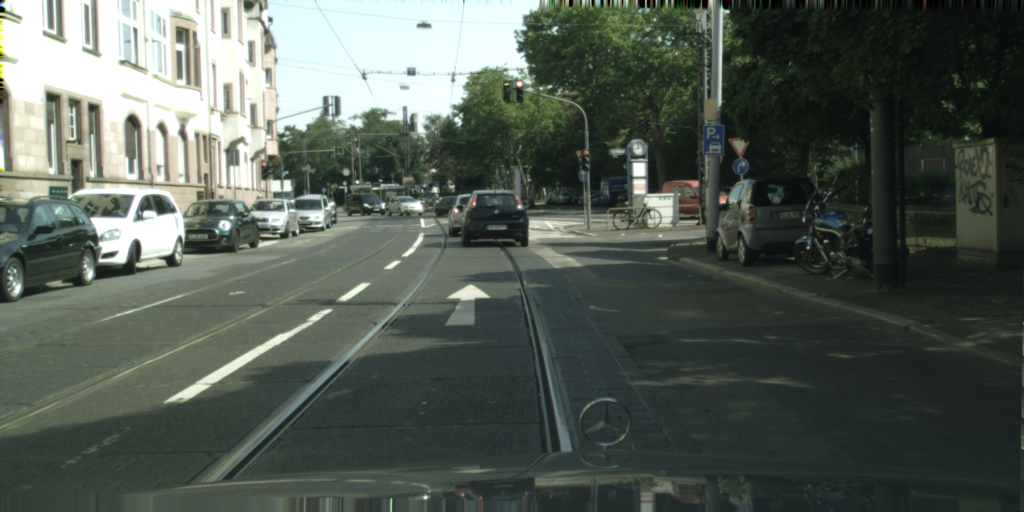}&
	\includegraphics[width=\imwidth]{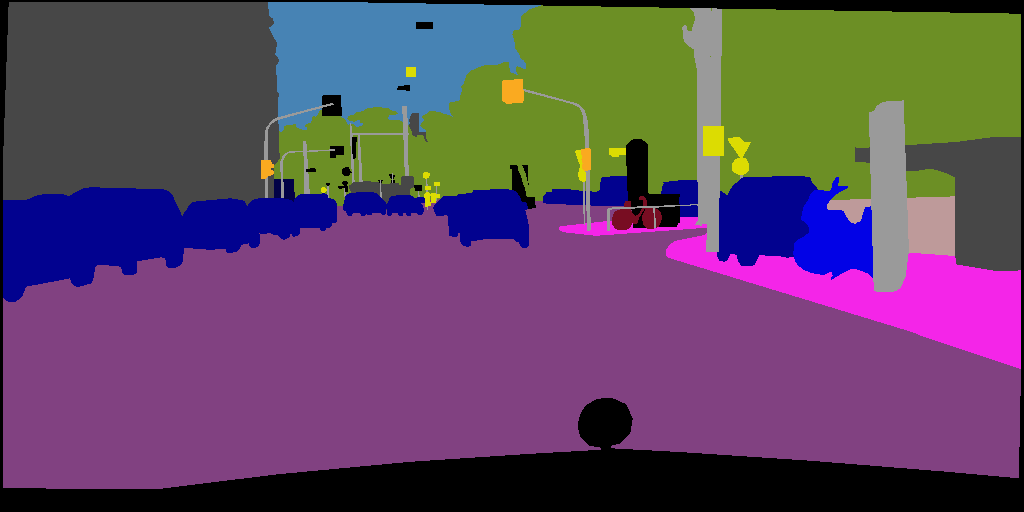}&
	\includegraphics[width=\imwidth]{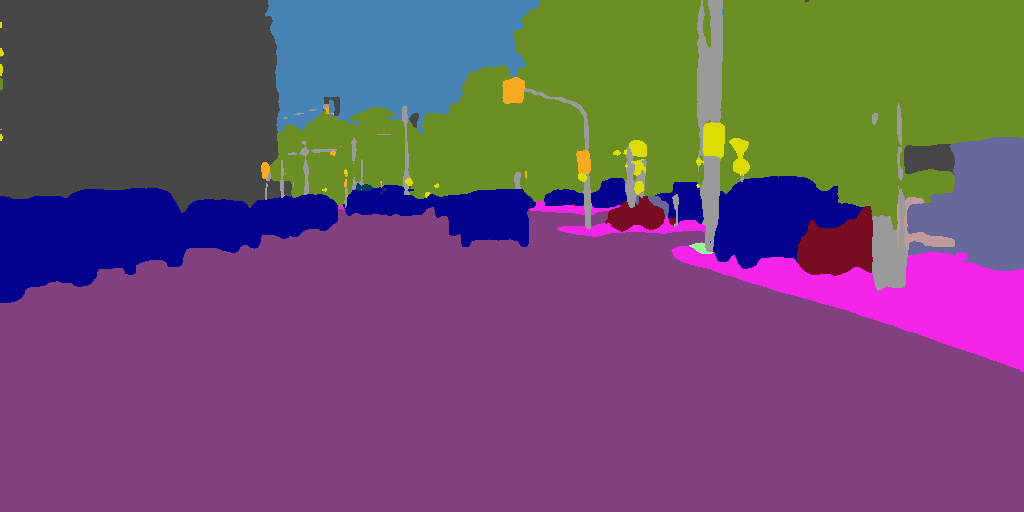}&
	\includegraphics[width=\imwidth]{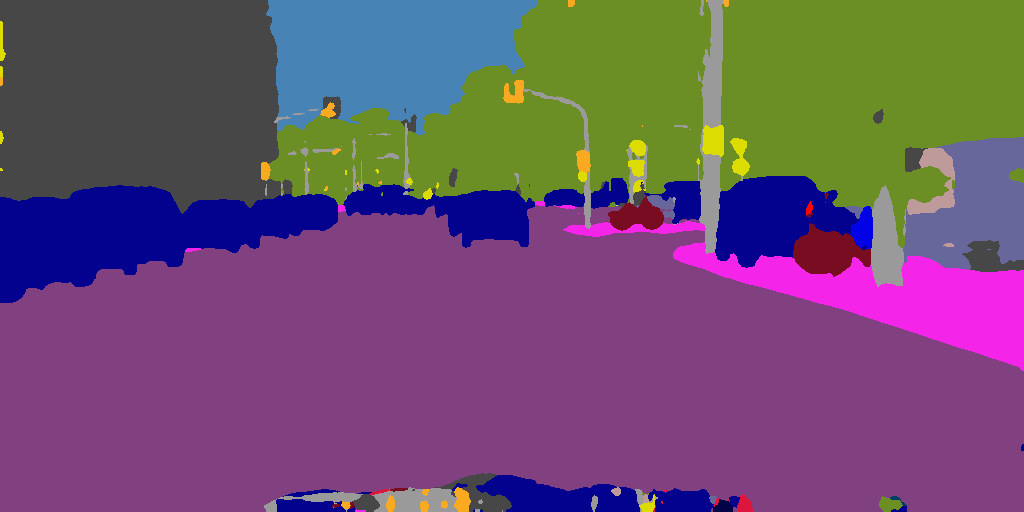}\\
	\includegraphics[width=\imwidth]{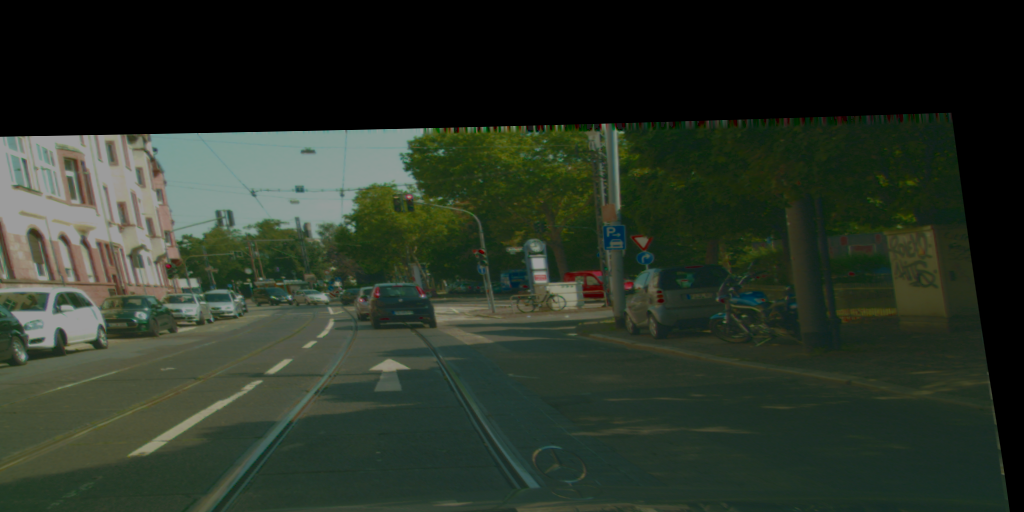}&
	\includegraphics[width=\imwidth]{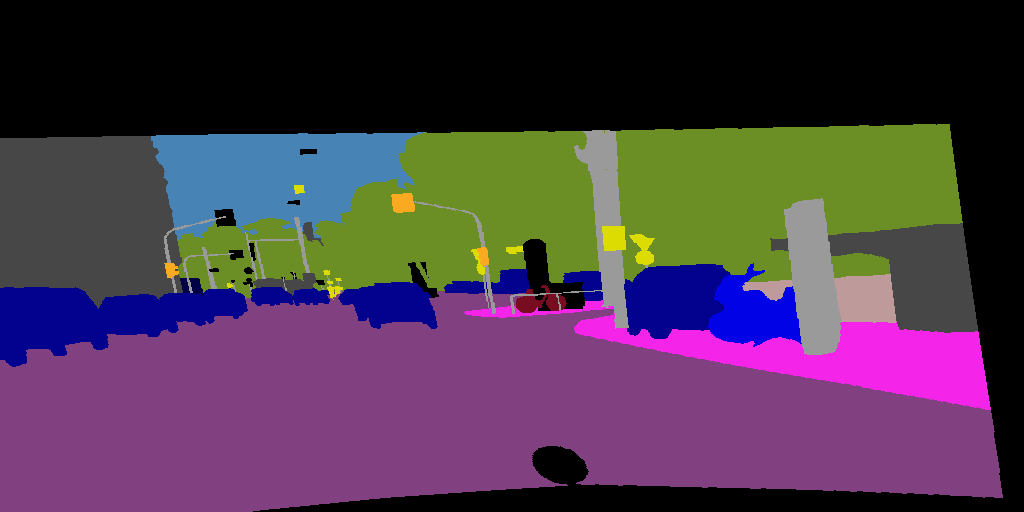}&
	\includegraphics[width=\imwidth]{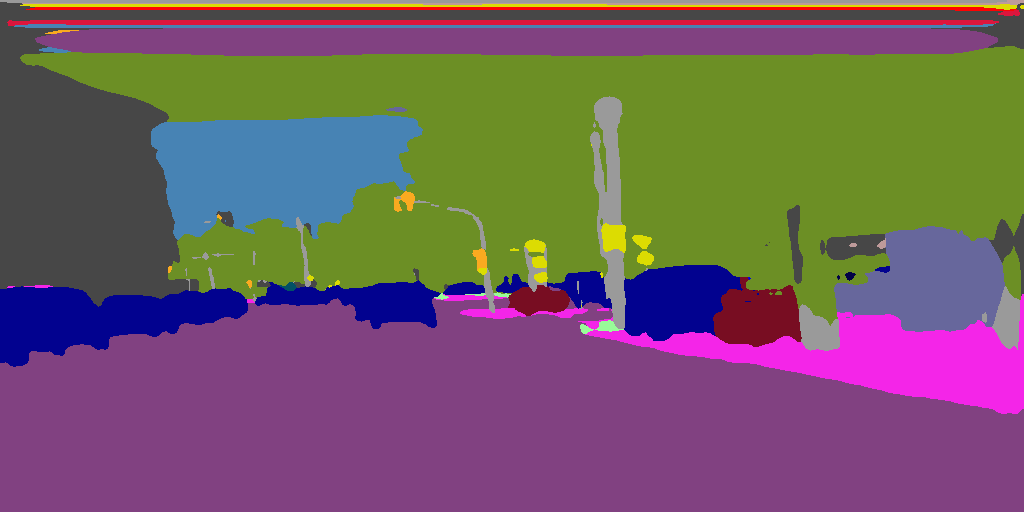}&
	\includegraphics[width=\imwidth]{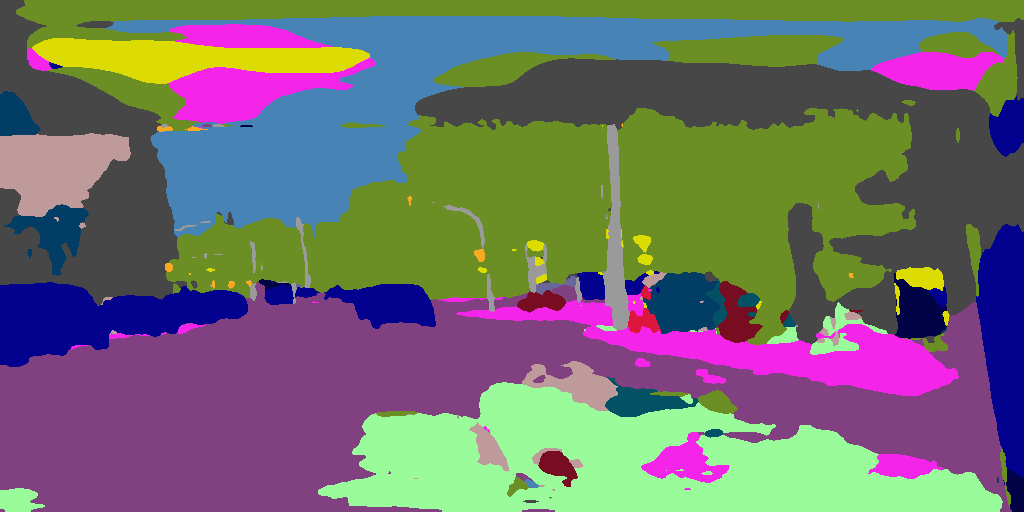}\\
	\includegraphics[width=\imwidth]{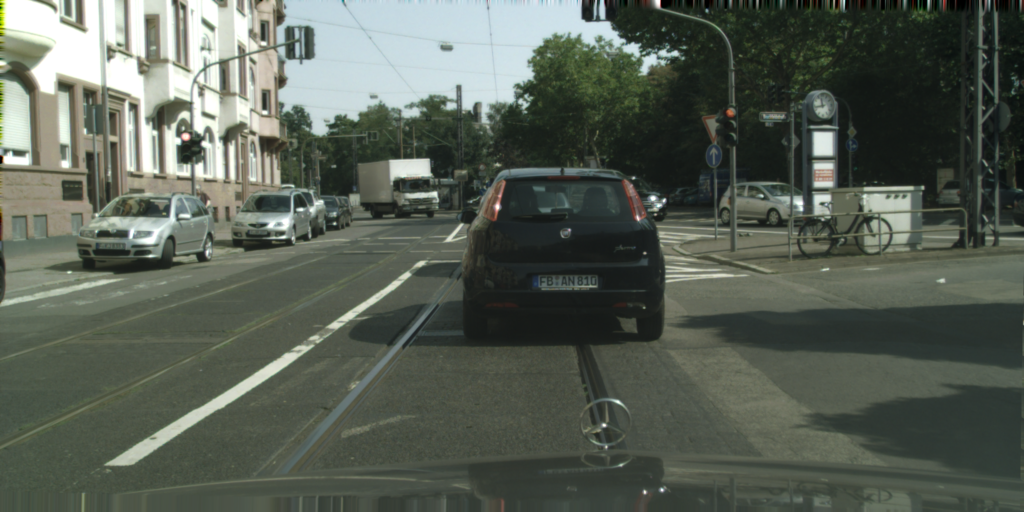}&
	\includegraphics[width=\imwidth]{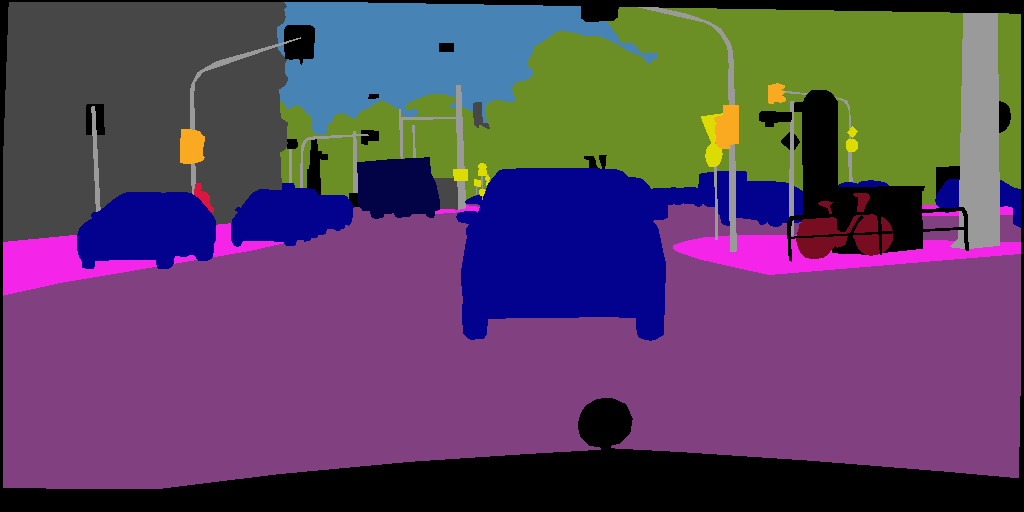}&
	\includegraphics[width=\imwidth]{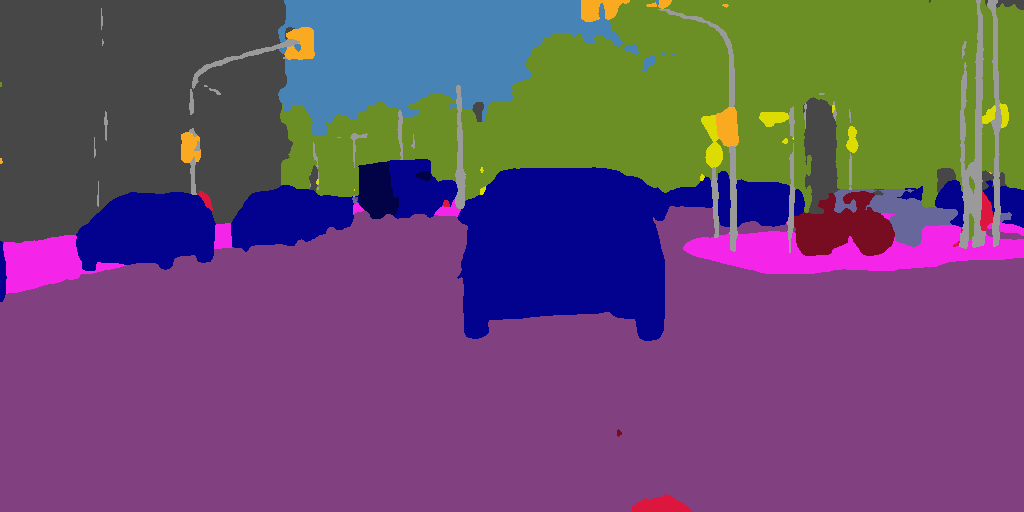}&
	\includegraphics[width=\imwidth]{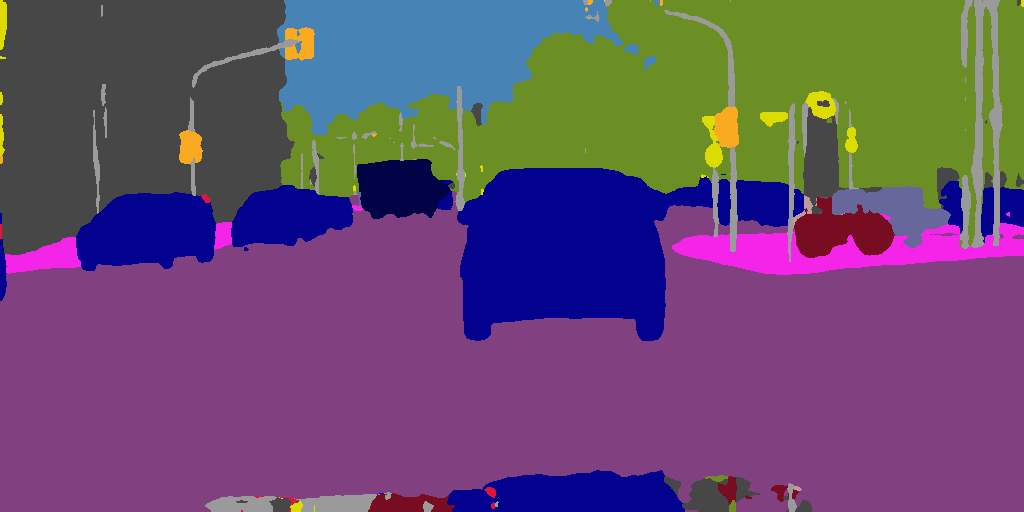}\\
	\includegraphics[width=\imwidth]{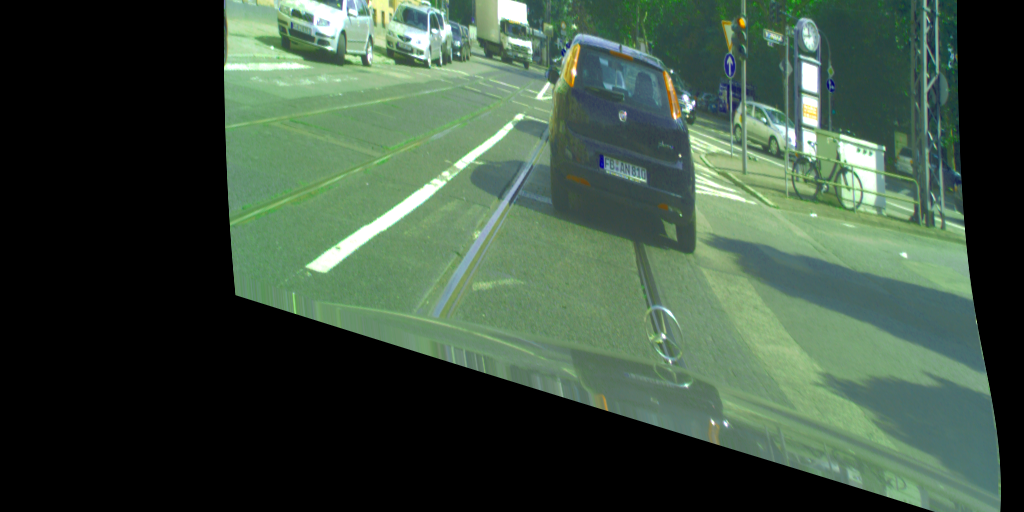}&
	\includegraphics[width=\imwidth]{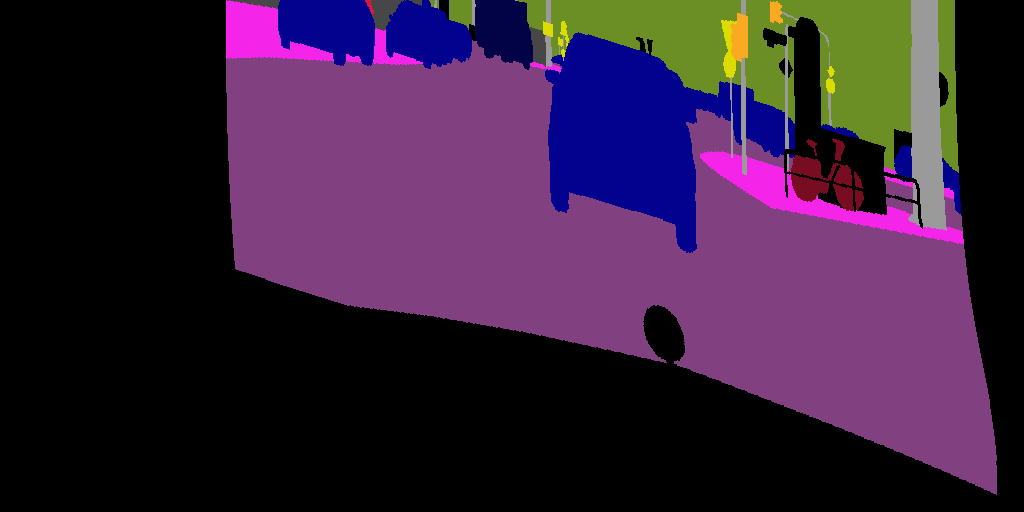}&
	\includegraphics[width=\imwidth]{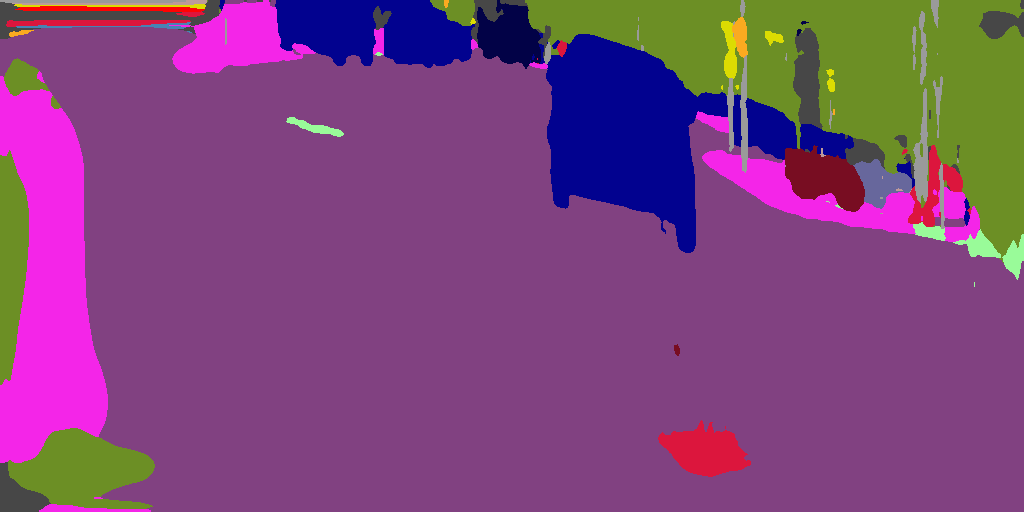}&
	\includegraphics[width=\imwidth]{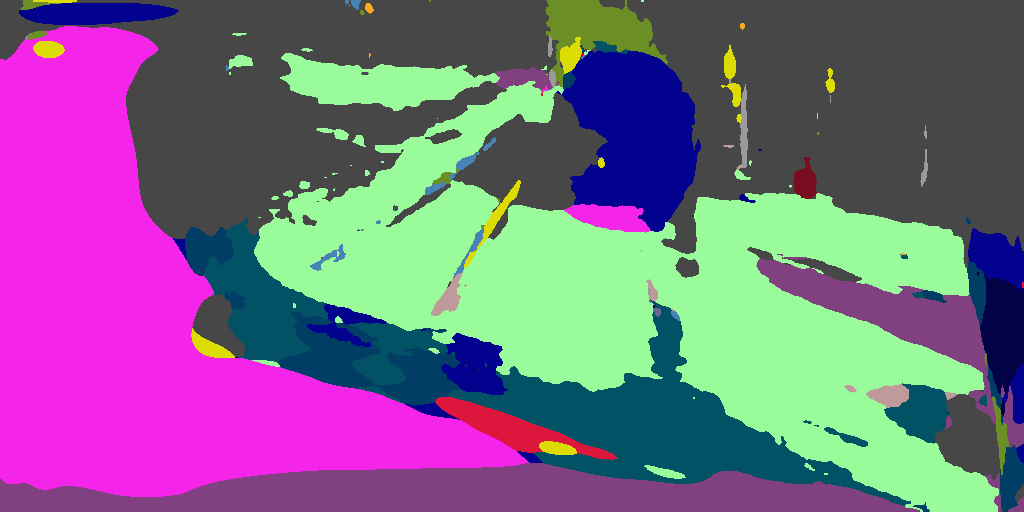}\\
	\includegraphics[width=\imwidth]{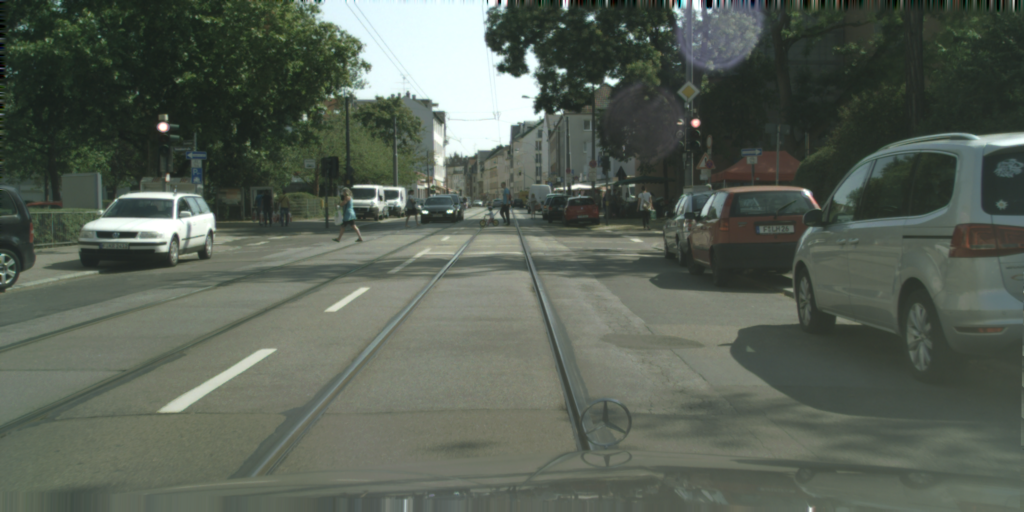}&
	\includegraphics[width=\imwidth]{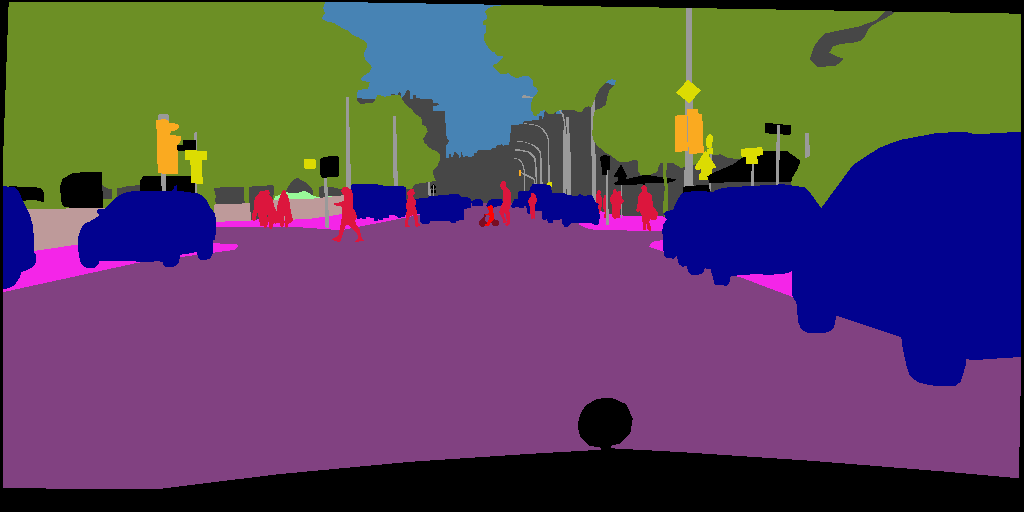}&
	\includegraphics[width=\imwidth]{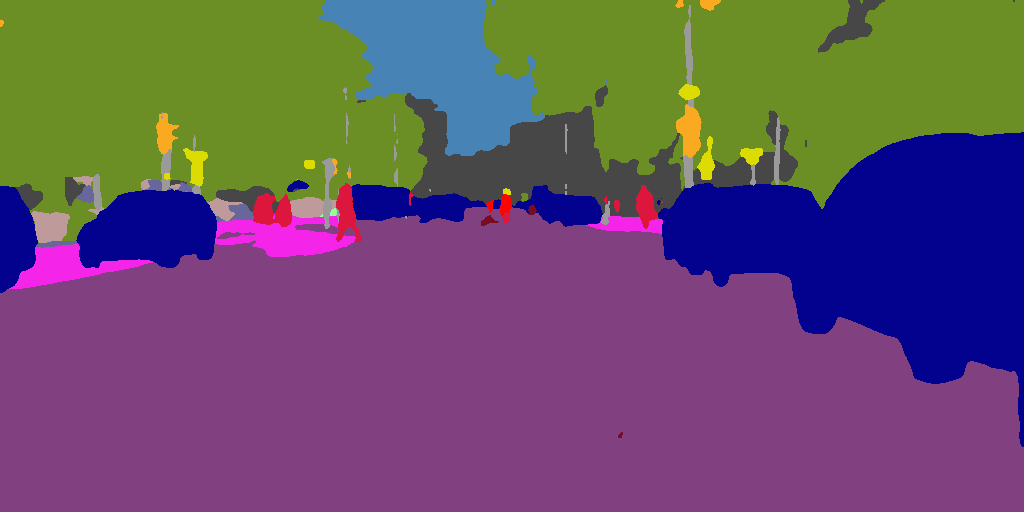}&
	\includegraphics[width=\imwidth]{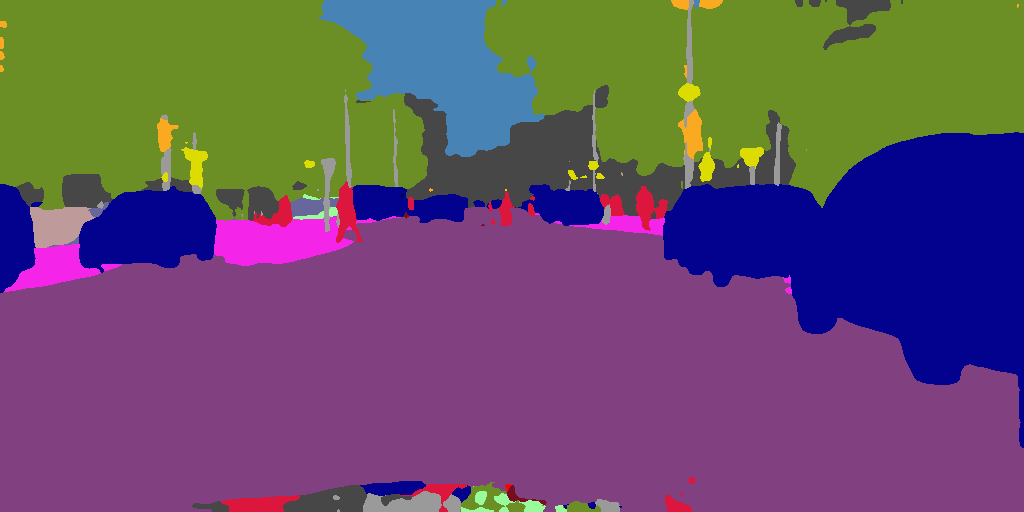}\\
	\includegraphics[width=\imwidth]{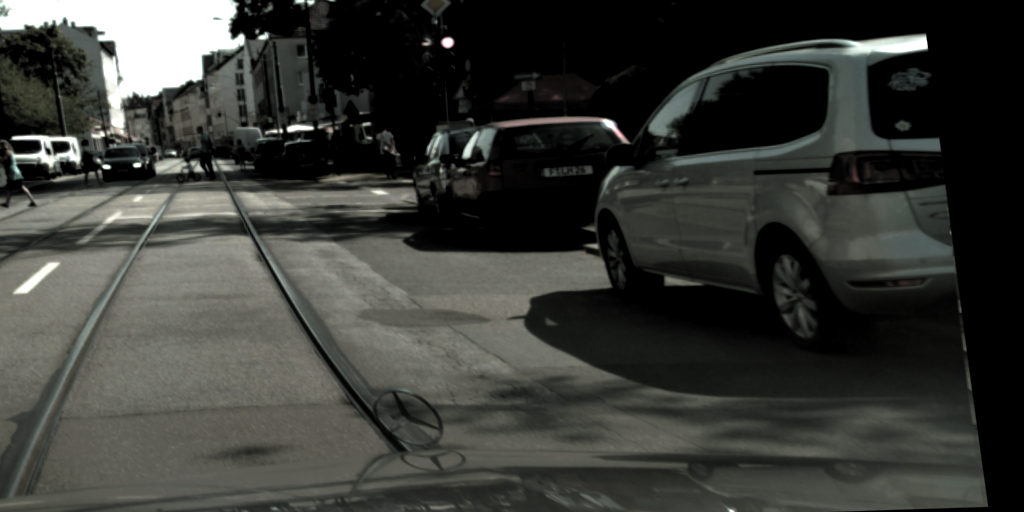}&
	\includegraphics[width=\imwidth]{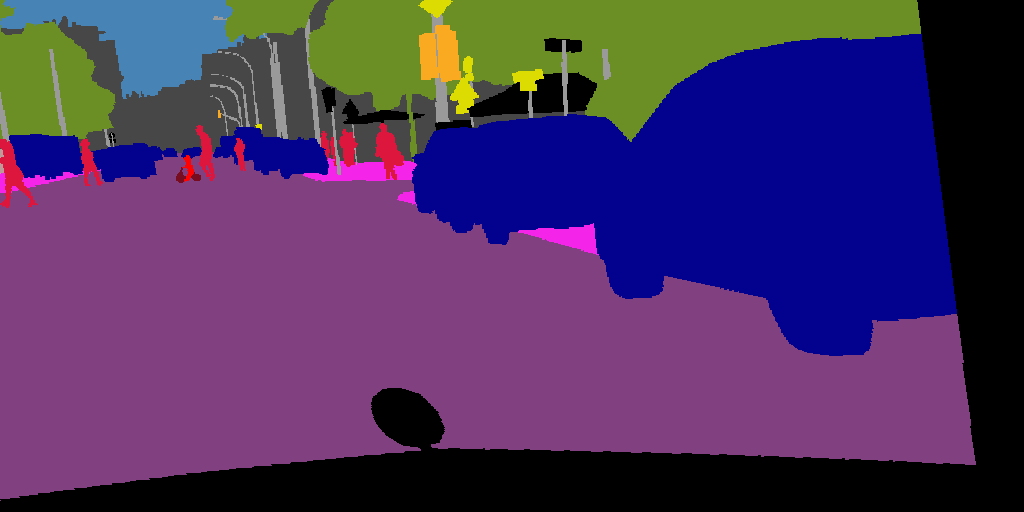}&
	\includegraphics[width=\imwidth]{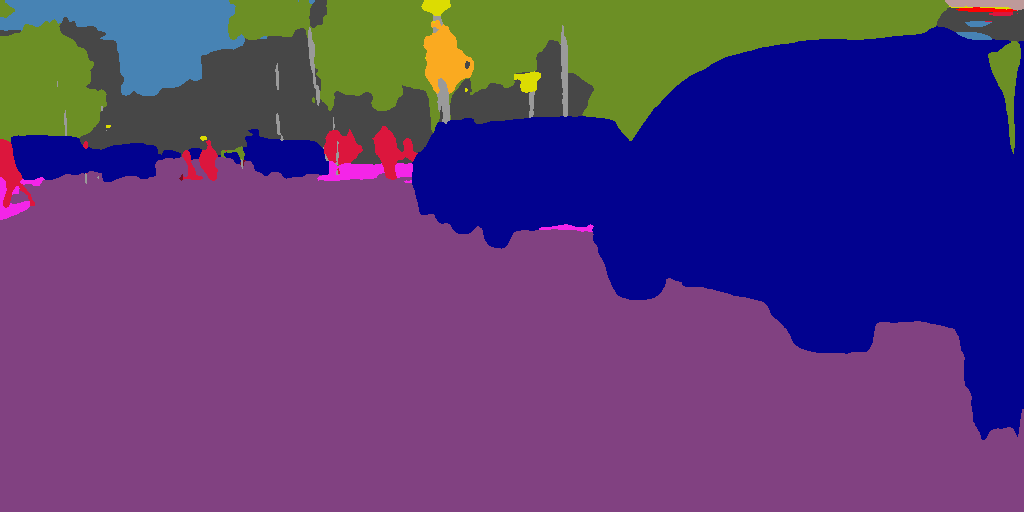}&
	\includegraphics[width=\imwidth]{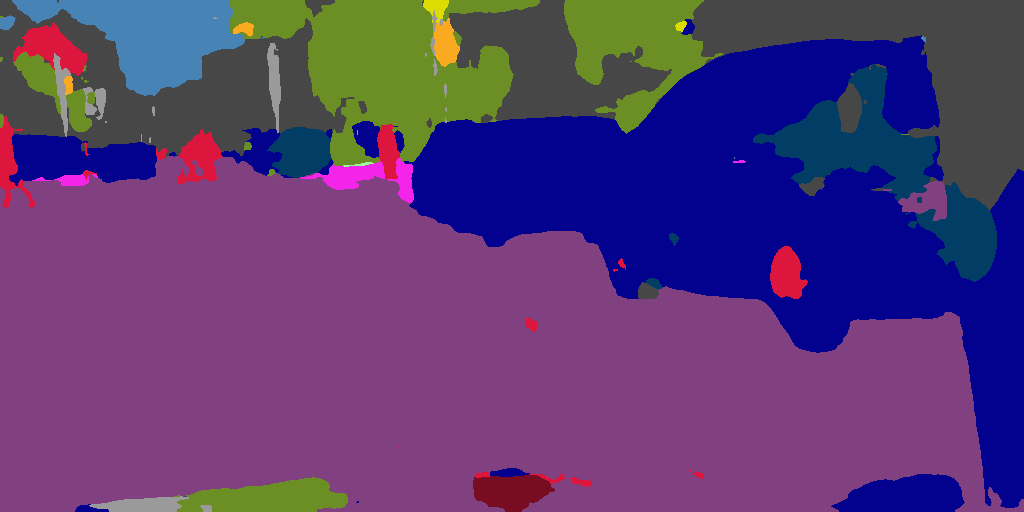}\\
	\includegraphics[width=\imwidth]{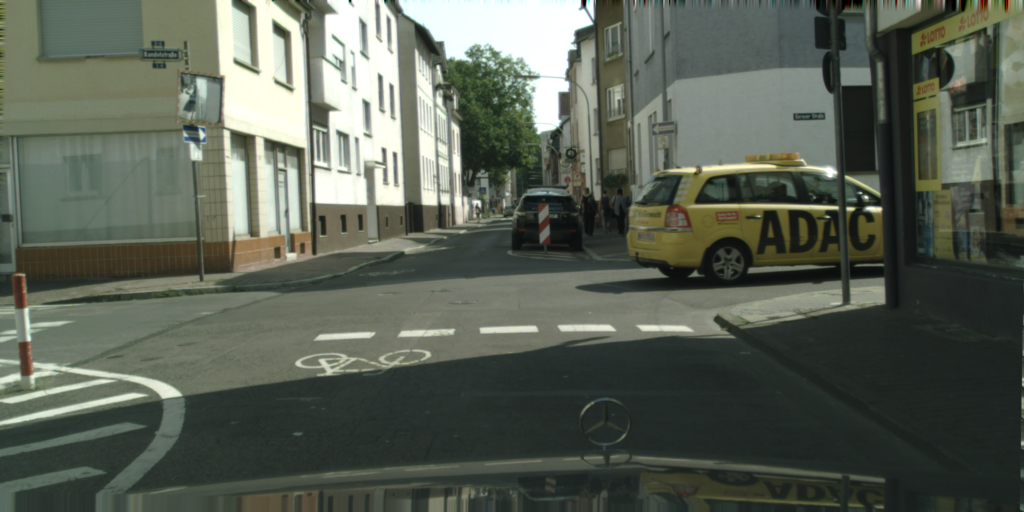}&
	\includegraphics[width=\imwidth]{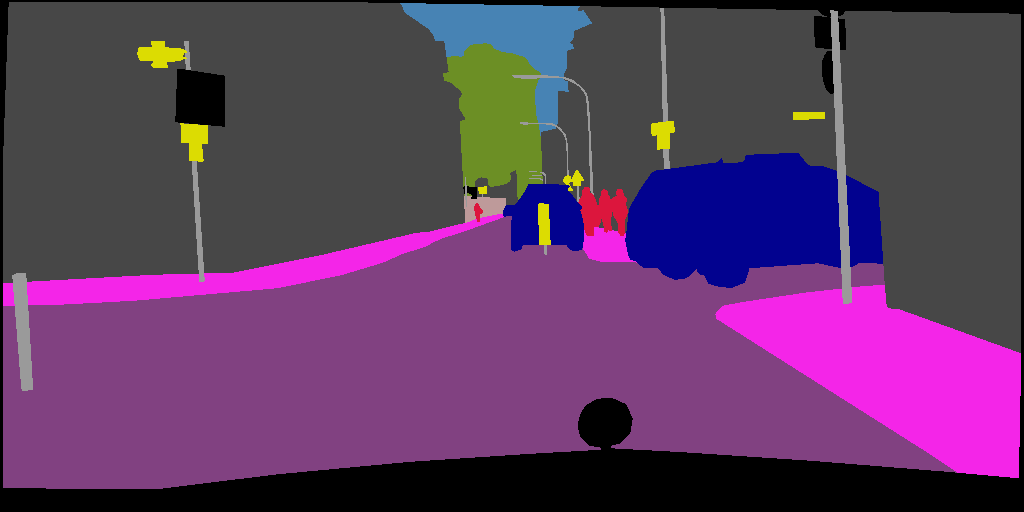}&
	\includegraphics[width=\imwidth]{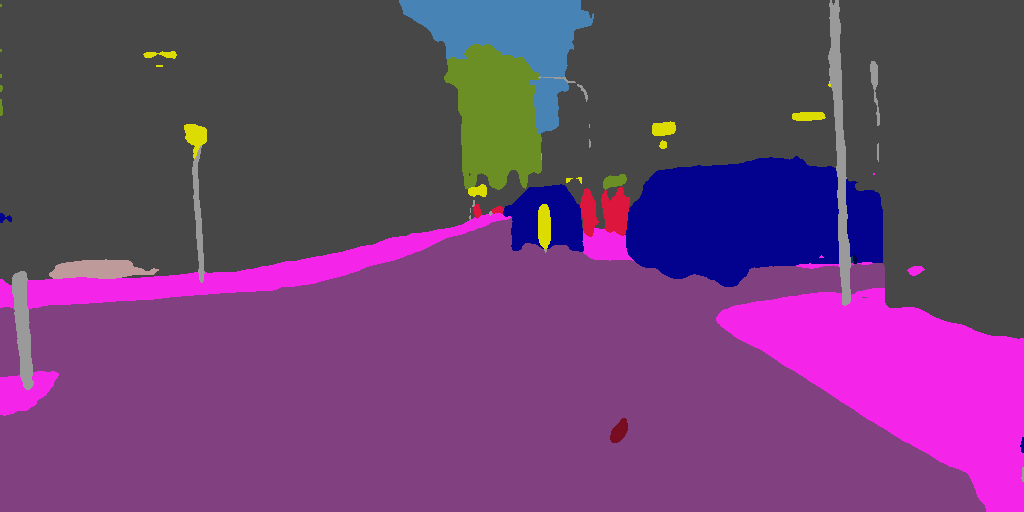}&
	\includegraphics[width=\imwidth]{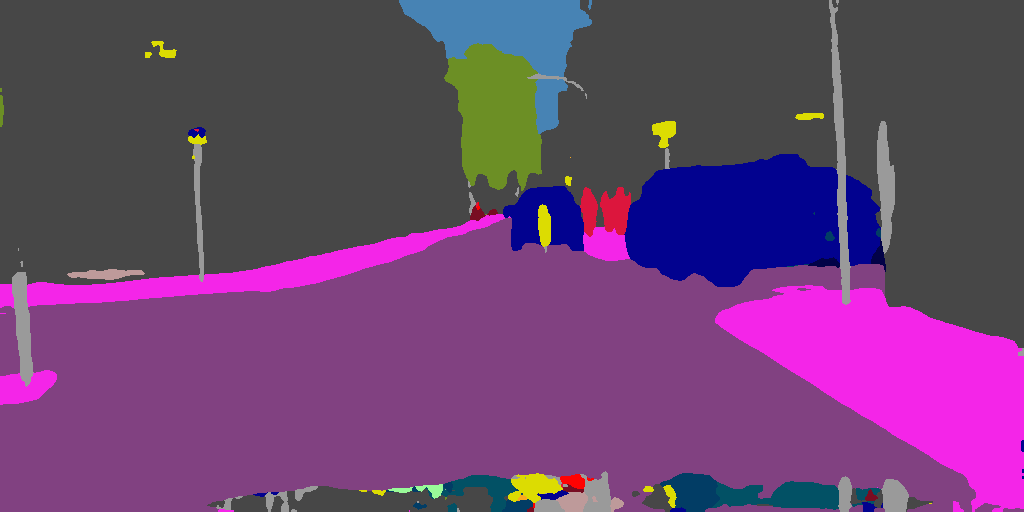}\\
	\includegraphics[width=\imwidth]{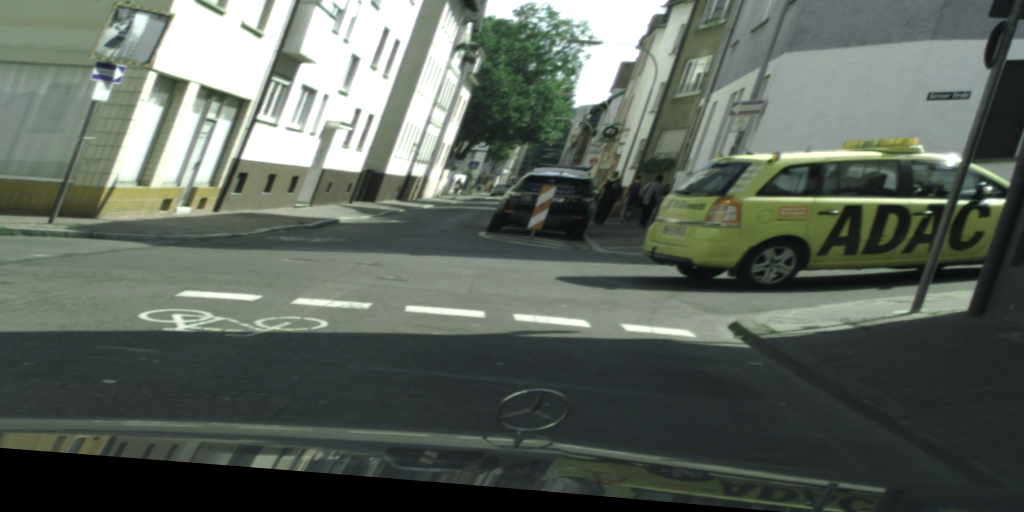}&
	\includegraphics[width=\imwidth]{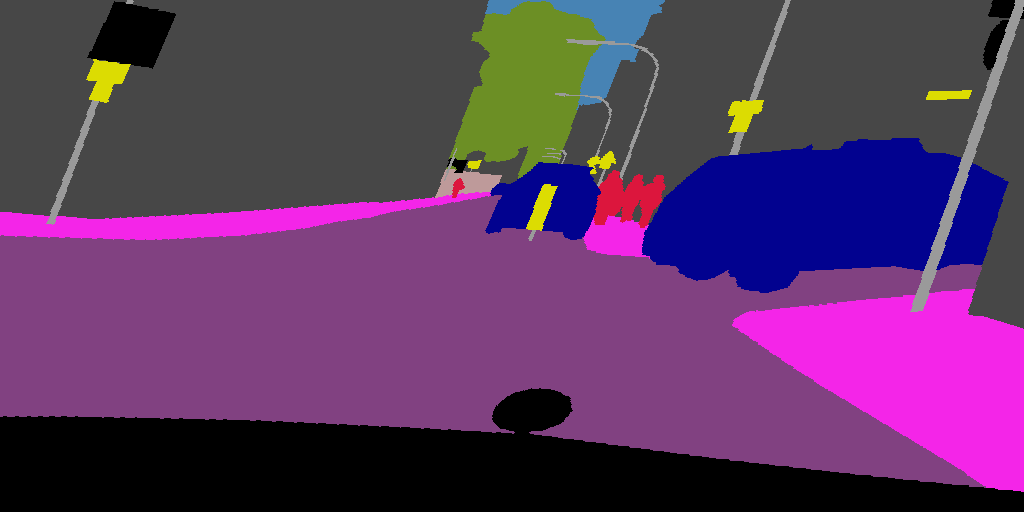}&
	\includegraphics[width=\imwidth]{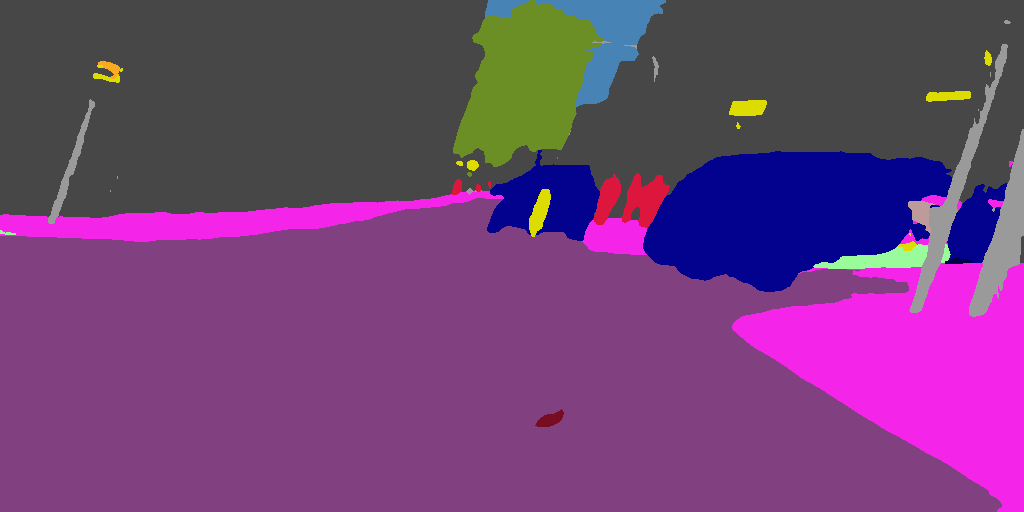}&
	\includegraphics[width=\imwidth]{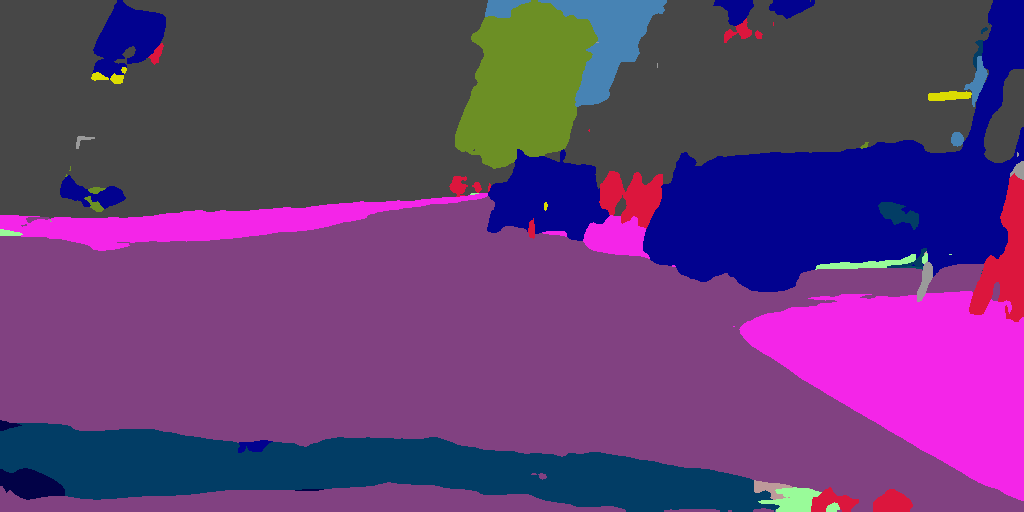}
\end{tabular}